\documentclass[reqno]{amsart}
\usepackage[utf8]{inputenc}
\usepackage{enumerate, graphicx, subcaption, amssymb, adjustbox,bbm}
\usepackage{tikz-cd}
\usepackage{url}
\usepackage{algorithm}
\usepackage{listings}
\usepackage{todonotes}
\usepackage{bbm}
\usepackage{soul}
\usepackage{dsfont}

\newcommand{\bD}{\mathbf{D}}
\newcommand{\bd}{\mathbf{d}}

\newcommand{\bX}{\mathbf{X}}
\newcommand{\bx}{\mathbf{x}}
\newcommand{\bw}{\mathbf{w}}
\newcommand{\bbeta}{\boldsymbol{\beta}}
\newcommand{\balpha}{\boldsymbol{\alpha}}
\newcommand{\btheta}{\boldsymbol{\theta}}

\newcommand{\bz}{{\mathbf{z}}}

\newcommand{\calF}{{\mathcal{F}}}

\newcommand{\E}{{\mathbb{E}}}
\renewcommand{\P}{{\mathbb{P}}}

\def\R{{\mathbb R}}  
\def\N{{\mathbb N}}  

\newtheorem{theo}{Theorem}

\theoremstyle{definition}

\newtheorem{remark}[theo]{Remark}

\author{M.\ Lindholm, R.\ Richman, A.\ Tsanakas, M.V.\ W{\"u}thrich}
\date{Version of \today}

\title[discrimination-free insurance pricing]{A multi-task network approach for calculating discrimination-free insurance prices}

\begin{document}

\begin{abstract}
In applications of predictive modeling, such as insurance pricing, indirect or proxy discrimination is an issue of major concern. Namely, there exists the possibility that protected policyholder characteristics are implicitly inferred from non-protected ones by predictive models, and are thus having an undesirable (or illegal) impact on prices. A technical solution to this problem relies on building a best-estimate model using all policyholder characteristics (including protected ones) and then averaging out the protected characteristics for calculating individual prices. However, such approaches require full knowledge of policyholders' protected characteristics, which may in itself be problematic. Here, we address this issue by using a multi-task neural network architecture for claim predictions, which can be trained using only partial information on protected characteristics, and it produces prices that are free from proxy discrimination. We demonstrate the use of the proposed model and we find that its predictive accuracy is comparable to a conventional feed-forward neural network (on full information). However, this multi-task network has clearly superior performance in the case of partially missing policyholder information. 
\end{abstract}

\maketitle



\section{Introduction}\label{sec: intro}

The question of avoiding discrimination in insurance pricing is becoming
increasingly important in many {markets} and jurisdictions. For example, the European Council \cite{EuropeanCouncil}
prohibits using gender information as a rating factor for insurance pricing; for an actuarial overview on discrimination
regulation we refer to Frees--Huang \cite{FreesHuang}. While regulation varies across jurisdictions, it is typically required that both direct and indirect discrimination be avoided. If $\bD$ denotes protected
information whose use is regarded as discriminatory, {\it direct discrimination} is avoided by merely 
not including $\bD$ in the regression model used for insurance pricing. However, not including protected information in the regression model is not necessarily sufficient for avoiding discrimination more broadly, because the protected information $\bD$ may
also be (implicitly) inferred from the non-protected variables, denoted by $\bX$. We call the impact of such inference on prices {\it indirect discrimination}. We note that this is a narrow use of the latter term, equivalent to what is also known as {\it proxy discrimination}, and does not consider any aspects of fairness and disparate impact;
we also refer to Prince--Schwarcz \cite{PrinceSchwarcz}, Frees--Huang \cite{FreesHuang},
Lindholm et al.~\cite{Lindholm}, Xin--Huang \cite{XinHuang} and Grari et al.~\cite{GrariEtAl} for relevant discussions.

In this paper, we address the problem of avoiding indirect discrimination in the calculation of insurance prices. Lindholm et al.~\cite{Lindholm} give a mathematical definition of direct and indirect
discrimination. Their approach for avoiding indirect discrimination
amounts to, first, using {\it all} available information
$(\bX, \bD)$ to calculate the
so-called {\it best-estimate price}. In a second step, one
removes the potential discriminatory dependence between $\bX$ and $\bD$ by marginalizing the best-estimate price 
w.r.t.~a pricing distribution which does not allow one to infer (or proxy) the protected
information $\bD$ from the non-protected variables $\bX$. 
This step removes the statistical
dependence between the two sets of information and results in the so-called
{\it discrimination-free insurance price} as defined in Lindholm et al.~\cite{Lindholm}.
This removal of statistical dependence can be motivated (and justified) by concepts
of causal statistics, see Lindholm et al.~\cite{Lindholm} and Araiza Iturria et al.~\cite{AraizaIturriaEtAl}; for an antecedent of this approach in economics, see Pope--Sydnor \cite{PopeSydnor}.

An attractive feature of the above discrimination-free insurance pricing approach is that any pricing model can be used to obtain the best-estimate price, which is subsequently adjusted to remove the potential for $\bD$ to be proxied by $\bX$. Moreover, the suggested procedure ensures that all potential indirect discrimination is removed, where it exists, and this is achieved regardless of the ability of the particular class of regression model used to infer information about $\bD$ from $\bX$. In particular, there is no need to explicitly quantify the potential impact of indirect discrimination before applying the method.

Thus, the calculation of discrimination-free insurance prices can be carried out using
any reasonable pricing model, if one has access to the {\it full} covariate information $(\bX, \bD)$. 
In practice, however, one may assume that the protected characteristics $\bD$ contain covariates that are considered sensitive, such as, e.g., ethnicity. Then, it will generally not be feasible to collect this information for all insurance contracts in the portfolio. As a consequence, it remains unclear how 
a model for discrimination-free insurance prices should be fitted, when discriminatory information is incomplete.
The goal of this paper is to address precisely this issue.
We present a {\it multi-output neural network} for {\it multi-task learning}, i.e., the proposed
network architecture performs simultaneously different regression tasks. 
This proposed network
architecture can also be fitted on incomplete protected information, and still provides
accurate results. That is, to fit our network architecture we only need the protected information $\bD$
on a part of the portfolio, but we can still receive a good predictive regression model for
discrimination-free insurance pricing. In particular, our proposal allows for more robust
fitting compared to just ignoring insurance policies with missing protected information. 

We illustrate the proposed methodology via a detailed case study, using a synthetic data set. The example demonstrates, first, that the multi-output network architecture provides results of comparable accuracy to a conventional feed-forward neural network, when complete information on policyholder characteristics is available. Second, when information on policyholder data is incomplete, the multi-task network greatly outperforms a conventional approach, whereby a regression model is only trained on those instances for which the full information is available -- both in the case where data are missing at random and not at random. The superiority of the multi-task architecture is more prevalent in the more realistic case where the data drop-out of protected characteristics is high.

\medskip

{\bf Organization of manuscript.}
In the next section we review the framework of discrimination-free insurance
pricing as introduced in Lindholm et al.~\cite{Lindholm}.
Section \ref{Multi-output network regression model} presents our solution
to the problem of having incomplete protected information, by gradually building up towards the multi-task network architecture.
Section \ref{Synthetic health insurance example} provides the synthetic data
example, which verifies the superior predictive power of our proposal
against a naive way of just ignoring insurance policies with incomplete data.  
Finally, in Section \ref{Conclusions} we give some concluding remarks.


\section{Discrimination-free insurance pricing}\label{sec: discrimination-free pricing}
We first recall the mathematical definitions of the best-estimate, unawareness and 
discrimination-free insurance prices,
as they were introduced in Lindholm et al.~\cite{Lindholm}. 
Throughout, we work on a probability space $(\Omega, \calF, \P)$
that is assumed to be sufficiently rich to carry all the objects that
we would like to study, and $\P$ denotes the physical probability measure. Our goal
is to employ a regression model that calculates the prices of insurance policies that satisfy the property
of being discrimination-free according to Definition 12 of Lindholm et al.~\cite{Lindholm}.

We assume that the vector of covariates $(\bX,\bD)$ can be partitioned into
non-discriminatory covariates $\bX$ and discriminatory covariates 
(protected characteristics) $\bD$. This split
into $\bX$ and $\bD$ is given exogenously, e.g., by law or by societal norms and preferences.
The distribution of the covariates $(\bX,\bD)$ of a randomly selected
policyholder is described by $\P$. The (insurance) claim of this policyholder is denoted
by $Y$, and we assume that this claim depends on the covariates $(\bX,\bD)$. That is, 
we typically would like to study the conditional distribution function
\begin{equation*}
y \in \R \quad \mapsto \quad F_{Y|(\bx,\bd)}(y) = \P \left[Y \le y \mid  \bX=\bx,\bD=\bd \right],
\end{equation*}
of a  selected policyholder having covariates $(\bX,\bD)=(\bx,\bd)$.

The {\it best-estimate price} of the policyholder with covariates $(\bX,\bD)$ is defined by
the conditional expectation (subject to existence)
\begin{equation}\label{best-estimate price}
\mu(\bX, \bD) := \E\left[Y \mid \bX, \bD\right].
\end{equation}
This price is called best-estimate because, under square integrability, it minimizes the conditional mean squared error of prediction (MSEP), given full information $(\bX,\bD)$. Thus, the
best-estimate price \eqref{best-estimate price} is the most accurate price we can calculate
under full covariate information $(\bX,\bD)$.
The general statistical problem is to optimally determine (estimate) this regression function
\begin{equation}\label{full regression function}
(\bx, \bd)~\mapsto~
\mu(\bx, \bd)= \E\left[Y \mid \bX=\bx, \bD=\bd\right],
\end{equation}
from past data (and maybe expert opinion). 
Below, we are going to use a neural network regression approach for this task.

Obviously this best-estimate price \eqref{best-estimate price} (potentially) directly discriminates
because it uses the discriminatory covariates $\bD$ as input. This motivates the
definition of the {\it unawareness price}, ignoring any knowledge about the discriminatory
covariates $\bD$,
\begin{equation}\label{unawareness price}
\mu(\bX) := \E\left[Y \mid \bX\right].
\end{equation}
The unawareness price \eqref{unawareness price} avoids {\it direct discrimination} according
to Definition 10 of Lindholm et al.~\cite{Lindholm} because we no longer need any
information about the discriminatory covariates $\bD$ to calculate this price.
Using the tower property of conditional expectations, we can rewrite the
unawareness price as follows 
\begin{equation}\label{tower property}
\mu(\bX) = \int_{\bd} \mu\left(\bX, \bd\right) ~{\rm d}\P(\bd \mid \bX),
\end{equation}
where $\P(\bd \mid \bX)$ describes the conditional distribution of the discriminatory
covariates $\bD$, given the non-discriminatory information $\bX$. It is exactly this link which is 
problematic, namely, having broad non-discriminatory information $\bX$ may (easily) allow
us to infer the discriminatory information $\bD$. Such an inference of the protected characteristics $\bD$ is 
therefore implicit in the definition of the unawareness price. This implicit inference has been
coined in insurance as
{\it proxy discrimination}, see, e.g., Frees--Huang \cite{FreesHuang} and Xin--Huang \cite{XinHuang}, or {\it indirect discrimination}, see Lindolm et al.~\cite{Lindholm}. To prevent indirect
discrimination, one needs to break the link that allows one to infer $\bD$ from $\bX$. This
can be done purely statistically by just replacing the outer distribution in \eqref{tower property}
by an unconditional one.
This replacement can be justified by arguments from causal statistics if insurance claims follow
a certain causal relationship, see Lindholm et al.~\cite{Lindholm} and
Araiza Iturria et al.~\cite{AraizaIturriaEtAl}.

These arguments motivate the definition of the {\it discrimination-free insurance price}
\begin{equation}\label{discrimination-free price}
\mu^*(\bX) := \int_\bd \mu (\bX, \bd)~{\rm d}\P^*(\bd),
\end{equation}
where the pricing distribution $\P^*(\bd)$ is dominated by the marginal distribution of the discriminatory
covariates $\bD\sim \P(\bd)$.

Discrimination-free insurance pricing \eqref{discrimination-free price}
has two ingredients, namely, the regression function $\mu(\bx,\bd)$, see \eqref{full regression function},
and the pricing distribution $\P^*(\bd)$. The most natural choice for this pricing
distribution is simply the marginal distribution $\P(\bd)$, but there may be other (justified)
choices, e.g., providing unbiasedness of discrimination-free insurance prices; for a broader discussion
on the choice of $\P^*$
we refer to Remark 7 and Section 4 in Lindholm et al.~\cite{Lindholm}.

In this paper we are more concerned about the first issue, namely, about 
selecting, estimating and applying the (best-estimate) regression function $(\bx,\bd)\mapsto\mu(\bx,\bd)$. 
In practice, this requires that we hold
both non-discriminatory {\it and} discriminatory information $(\bx,\bd)$
from the insurance policyholders for regression model fitting,
and the discriminatory information is integrated out (adjusted for) only in the subsequent (second) step
\eqref{discrimination-free price}.
However, in many cases it is problematic to collect this discriminatory information over the entire
insurance portfolio.
Therefore, fitting the regression function \eqref{discrimination-free price} might not be practical.
In the next section, we provide a technical workaround which requires discriminatory
information only for part of the portfolio, but it will still equip us with accurate predictive models.

\begin{remark}
\label{demographic discrimination}
The discrimination-free insurance price \eqref{discrimination-free price} is defined
within a given model specification, i.e., for a given distributional model for $(Y,\bX,\bD)$,
see Definition 12 of Lindholm et al.~\cite{Lindholm}. This does not consider model error
coming from a poorly specified stochastic model for $(Y,\bX,\bD)$, which
may result in a different form of discrimination, e.g., arising from
a certain sub-population being under-represented in the data. Naturally, on the corresponding part
of the covariate space we have greater model uncertainty, because we have less data for
an accurate model fit. This may result in forms of demographic discrimination that are outside
of our (more narrow) scope which is always attached to a {\it given}
stochastic model. For a discussion of discrimination arising from unrepresentative data in a different
context, see Buolamwini--Gebru \cite{BuolamwiniGebru}.
\end{remark}


\section{Multi-output network regression model}
\label{Multi-output network regression model}
We present statistical modeling of the regression function $\mu(\bx,\bd)$
within the framework of feed-forward neural
networks (FNNs). We start by introducing a standard (plain-vanilla) FNN
architecture, and in a second step we discuss how this FNN architecture can be modified to serve our purpose of deriving discrimination-free insurance prices with partial information of the protected
characteristics.
The notation and terminology of neural network regression modeling is taken from
W\"uthrich--Merz \cite{WM2021}.

\subsection{Feed-forward neural network regression model}
Assume that the regression function can be modeled by a FNN architecture
taking the following form
\begin{equation}\label{plain-vanilla FNN}
(\bx, \bd)~\mapsto~
g \left(\mu(\bx, \bd)\right) = \left\langle \bbeta,  \bz^{(m:1)}(\bx, \bd) \right\rangle ,
\end{equation}
where $g:\R \to \R$ is a strictly monotone and smooth link function, 
$\bz^{(m:1)}:\R^{q_0}\to \R^{q_m}$ is a FNN of depth $m\in \N$,
and $\bbeta=(\beta_0,\ldots, \beta_{q_m})^\top \in \R^{q_m+1}$
is the readout parameter providing the scalar product on the right-hand side of
\eqref{plain-vanilla FNN}
\begin{equation*}
\left\langle \bbeta,  \bz^{(m:1)}(\bx, \bd) \right\rangle
= \beta_0 + \sum_{j=1}^{q_m} \beta_j \, z_j^{(m:1)}(\bx, \bd).
\end{equation*}
The FNN $\bz^{(m:1)}$ of depth $m \in  \N$ is a composition of $m$ hidden FNN layers 
$\bz^{(j)}:\R^{q_{j-1}}\to \R^{q_j}$, $1\le j \le m$, providing
\begin{equation*}
(\bx, \bd)\in \R^{q_0}\quad \mapsto\quad 
  \bz^{(m:1)}(\bx, \bd)= \left(\bz^{(m)}\circ \ldots  \circ\,\bz^{(1)}\right) (\bx, \bd).
\end{equation*}
This maps the $q_0$-dimensional vector-valued input $(\bx, \bd)\in \R^{q_0}$ to a new
(learned) $q_m$-dimensional representation $\bz^{(m:1)}(\bx, \bd) \in \R^{q_m}$
of the original non-discriminatory and discriminatory covariates $(\bx, \bd)$.

Summarizing, FNN regression modeling requires specification of the network architecture.
This involves the
depth $m \in \N$, the number of neurons $q_j \in \N$ in each hidden layer $1\le j \le m$,
the activation function in each of these neurons, as well as the link function $g$. Such a network
architecture can then be fitted (trained) to the available data, which means that the
readout parameter $\bbeta$ as well as all parameters in the hidden
layers $1\le j \le m$ (called network weights $\bw$) are fitted to the available data. 
Successful FNN fitting involves in most cases an early
stopping strategy to prevent (in-sample) over-fitting the model to the training data, i.e., targeting
for an optimal out-of-sample predictive performance; for a detailed description of network fitting
we refer to Chapter 7 of W\"uthrich--Merz \cite{WM2021}.

This fitted FNN \eqref{plain-vanilla FNN} provides the best-estimate prices
\begin{equation}\label{plain vanilla best estimate}
\mu(\bx, \bd) = \E\left[Y \mid \bX=\bx, \bD=\bd\right]
= g^{-1}\left\langle \bbeta,  \bz^{(m:1)}(\bx, \bd) \right\rangle,
\end{equation}
for the insurance claims $Y$, being described by the covariates $(\bx, \bd)$.
From this we can calculate the discrimination-free insurance prices with formula
\eqref{discrimination-free price} by specifying a suitable pricing
distribution $\P^*(\bd)$. A standard choice is to use the marginal distribution of $\bD$
from the part of the portfolio where the protected information of $\bD$ is known.

The difficulty in practice with this approach, and similar regression approaches
such as generalized linear models (GLMs), is that it requires full knowledge
of the discriminatory information $\bD=\bd$ of the policyholders. Otherwise one cannot fit 
this FNN \eqref{plain-vanilla FNN} on the available (past) data. A naive solution is
to just fit this FNN architecture on the sub-portfolio where $\bD$ is available.
We call this the (naive) {\it plain-vanilla FNN approach} because it is clearly non-optimal
to disregard any insurance policy  where there is no complete information about
the covariates $(\bX,\bD)$ available.

\begin{remark} In the introduction we have mentioned that discrimination-free insurance
  pricing can be applied to any pricing (regression) model. Here, we restrict to FNN architectures
  which seems rather limiting. However, we would like to mention that large FNN architectures
  provide the universal approximation property. This implies that within FNN architectures
  we can mimick any other (sufficiently regular) regression model.
 \end{remark}

 \subsection{Multi-output neural network regression model}
 In constructing (and fitting) the plain-vanilla FNN best-estimate prices \eqref{plain vanilla best estimate},
we directly use the discriminatory information $\bD=\bd$ of the policyholders
 as an input variable to the FNN. Our proposal is to change
 this FNN architecture such that only the non-discriminatory information $\bX=\bx$ is used as an input variable
 to the network, but at the same time we generate a whole family of best-estimate prices that reflects
 the different specifications (levels) of the discriminatory information. In the present section
 we introduce this new network architecture. We call it a {\it multi-output FNN} architecture because it
 has multiple outputs that generate the family of models. In Section 
 \ref{Multi-task learning with multi-output neural networks}, below, we extend this multi-output FNN architecture
 to a {\it multi-task FNN} architecture which not only generates a whole family of models, but it also
 (internally) predicts the discriminatory information on the insurance policies where
 this information is missing. It will exactly be this
 multi-task FNN architecture that we promote for discrimination-free insurance pricing under incomplete
 discriminatory information, as it can deal with the issue of missing protected
 information, but still provides good predictive models. In this approach protected information
 will only be needed on part of the portfolio for model training.

Assume that the discriminatory information $\bD$ 
only takes finitely many values $\bd \in D_K:=\{\bd_1,\ldots, \bd_K\}$.
Typically, we think of discriminatory information being of categorical type, e.g., gender
or ethnicity.
If this is not the case, discriminatory information can be discretized, and one should
work with this discretized version.
We modify the above plain-vanilla FNN \eqref{plain-vanilla FNN}
such that it only considers non-discriminatory
covariates $\bx$ as an input giving us the
learned representation
\begin{equation*}
\bx \quad \mapsto\quad 
  \bz^{(m:1)}(\bx)= \left(\bz^{(m)}\circ \ldots  \circ\,\bz^{(1)}\right) (\bx).
\end{equation*}
This learned representation should be sufficiently rich such that it provides
a whole family of suitable regression functions, parametrized by $\bd_k \in D_K$.
 This typically requires that the number of
hidden neurons, in particular $q_m$ in the last hidden layer, is not too small.
This learned representation is now used to calculate the best-estimate prices simultaneously
for all discriminatory specifications $\bd_k \in  D_K$,
that is, we set for the {\it multi-output FNN} architecture
\begin{equation}\label{multi-output FNN}
\Big(\mu(\bx, \bd_1), 
\ldots, \mu(\bx, \bd_K)\Big) = \left(g^{-1}\left\langle \bbeta_1,  \bz^{(m:1)}(\bx) \right\rangle ,
\ldots, g^{-1}\left\langle \bbeta_K,  \bz^{(m:1)}(\bx) \right\rangle \right).
\end{equation}
In other words, for every non-discriminatory input $\bx$ we generate a whole
family of outputs that simultaneously provide the best-estimate prices $\mu(\bx, \bd_k)$ for all
levels $\bd_k \in D_K$, $1\le k \le K$, of the discriminatory information.
The different specifications $\bd_k \in D_K$ of the
discriminatory covariates $\bD$ are encoded in the readout parameters $\bbeta_1,\ldots, \bbeta_K
\in \R^{q_m+1}$,
and the only remaining question is about fitting these parameters (as well as the network weights $\bw$
in the hidden layers) to the available data.

We start by describing how the plain-vanilla FNN \eqref{plain-vanilla FNN} is fitted
to i.i.d.~data $(Y_i, \bX_i, \bD_i)_{1\le i \le n}$ following the same model.
For the moment we assume to be in the situation of complete information.
We choose a loss function $L:\R \times \R \to \R_+$ to assess the quality of a
fit. This loss function can be the square loss function, a deviance loss function, or any
other sensible choice that fits to the estimation problem to be solved.
If $\btheta=(\bbeta, \bw)$ collects all parameters to be estimated/fitted, then, typically, an optimal
parameter is found by solving (M-estimation)
\begin{equation}\label{plain vanilla fitting}
\widehat{\btheta} ~=~ \underset{\btheta}{\arg \min} \, \sum_{i=1}^n L \Big(Y_i, \,
\mu_{\btheta}(\bX_i,\bD_i) \Big),
\end{equation}
where in the regression function $\mu(\cdot)=\mu_{\btheta}(\cdot)$ we highlight its dependence
on the parameter $\btheta$ to be optimized. This is the process to fit
the plain-vanilla FNN given in \eqref{plain vanilla best estimate}, subject to early stopping
to prevent from in-sample over-fitting; for a detailed discussion of FNN
fitting we refer to Section 7.2.3 in W\"uthrich--Merz \cite{WM2021}. 

For fitting the multi-output FNN \eqref{multi-output FNN} we modify this fitting procedure as follows
\begin{eqnarray}\label{multi-output fitting}
\widehat{\btheta} &=& \underset{\btheta}{\arg \min} \, \sum_{i=1}^n 
\sum_{k=1}^K L \Big(Y_i, \,
\mu_{\btheta}(\bX_i, \bd_k) \Big)\, \mathds{1}_{\{\bD_i = \bd_k\}}
\\&=&
\underset{\btheta}{\arg \min} \, \sum_{i=1}^n 
\sum_{k=1}^K L \left(Y_i, \,
g^{-1}\left\langle \bbeta_k,  \bz^{(m:1)}(\bX_i) \right\rangle \right) \mathds{1}_{\{\bD_i = \bd_k\}},
\nonumber
\end{eqnarray}
where $\btheta=(\bbeta_1,\ldots, \bbeta_K, \bw)$ collects all readout parameters
and the network weights, see \eqref{multi-output FNN}.
That is, we add an indicator $\mathds{1}_{\{\bD_i = \bd_k\}}$ referring to the
discriminatory information $\bD_i$ of observation $i$, which, in turn, trains the corresponding readout parameter $\bbeta_k \in \R^{q_m+1}$ of the multi-output FNN \eqref{multi-output FNN}. 
Note that this is
the {\it only} step where the protected information $\bD_i$ is used in the
multi-output FNN approach.

Having the fitted multi-output FNN \eqref{multi-output FNN} we arrive
at the discrimination-free insurance prices
\begin{equation}\label{DFIP discrete}
\bx ~\mapsto~
\mu^*(\bx) ~=~ \sum_{k=1}^K \mu (\bx, \bd_k)\,\P^*(\bd_k)
~=~ \sum_{k=1}^K g^{-1}\left\langle \bbeta_k,  \bz^{(m:1)}(\bx) \right\rangle\,\P^*(\bd_k),
\end{equation}
for each possible choice of the pricing distribution $\P^*$ on the finite set $D_K=\{\bd_1,\ldots, \bd_K\}$.

\medskip

To conclude, 
the multi-output FNN \eqref{multi-output FNN} generates a whole family of best-estimate
prices $(\mu(\cdot, \bd_k))_{1\le k \le K}$ of the discriminatory information in $D_K$. This discriminatory information only enters the loss function in the fitting procedure \eqref{multi-output fitting}, and once
this model is fitted we no longer need discriminatory information to calculate
the discrimination-free insurance price \eqref{DFIP discrete} for any (new) insurance policy.

\subsection{Multi-task learning and incomplete discriminatory information}
\label{Multi-task learning with multi-output neural networks}
We extend the multi-output FNN architecture from the previous section to a multi-task learning
model. This will be especially suitable for our problem of incomplete discriminatory information, because,
as a second task, this extended multi-output FNN will predict the discriminatory information on the policies
where this information is not available.

We define the categorical probabilities for $\bd_k \in D_K$, $1\le k \le K$,
\begin{equation*}
  p_k(\bx):=\P[\bD=\bd_k \mid \bX=\bx]
 ~ \in~ [0,1].
\end{equation*}
We choose two separate FNNs for representation learning 
\begin{equation}\label{separate FNNs}
\bx~\mapsto~
  \bz_\mu^{(m:1)}(\bx)\in \R^{q_m} \qquad \text{ and } \qquad 
  \bx ~\mapsto~ 
  \bz_p^{(m:1)}(\bx)\in \R^{q_m}.
\end{equation}
For simplicity we assume that these two FNNs have exactly the same network architecture,
but, typically, their network weights (parameters) $\bw_\mu$ and $\bw_p$ differ. The first
FNN is used to model the best-estimate prices 
\begin{equation}
\label{best estimates A}
\bx~\mapsto~
\mu(\bx, \bd_k)=g^{-1}\left\langle \bbeta_k,  \bz_\mu^{(m:1)}(\bx) \right\rangle
\qquad \text{ for $1\le k \le K$,}
\end{equation}
and with readout parameters $\bbeta_k \in \R^{q_m+1}$.
The second FNN is used to model the categorical probabilities $(p_k(\bx))_{1\le k \le K}$.
Using the softmax output function and for readout parameters $\balpha_1,
\ldots, \balpha_K \in \R^{q_m+1}$ we receive the FNN classification model
\begin{equation}
\label{best estimates B}
\bx~\mapsto~
p_k(\bx) = \frac{ \exp \left\langle \balpha_k,  \bz_p^{(m:1)}(\bx) \right\rangle}{ \sum_{j=1}^K
\exp \left\langle \balpha_j,  \bz_p^{(m:1)}(\bx) \right\rangle}~\in~(0,1).
\end{equation}
At the current stage these two networks are completely unrelated because they run
in parallel, and they can be fitted independently from each other. We now make them
related by (internally)
calculating the unawareness price using the tower property \eqref{tower property}, i.e.,
\begin{equation}
\label{best estimates C}
\bx~\mapsto~
\mu(\bx) = \sum_{k=1}^K \mu(\bx, \bd_k)\,p_k(\bx).
\end{equation}
Combining \eqref{best estimates A}, \eqref{best estimates B} and \eqref{best estimates C} we receive the {\it multi-task FNN} architecture
\begin{equation}\label{multi-task learning}
\bx ~ \mapsto ~
\Big(\mu(\bx, \bd_1), 
\ldots, \mu(\bx, \bd_K);\, 
p_1(\bx), \ldots, p_K(\bx);\,
\mu(\bx)\Big) ~\in~\R^{2K+1}, 
\end{equation}
with network parameter $\btheta=(\bbeta_1,\ldots, \bbeta_K,\balpha_1,\ldots, \balpha_K, \bw_\mu, \bw_p)$.
As input this multi-task FNN only uses the non-discriminatory information $\bx$. Remark that
the unawareness price $\mu(\bx)$ in \eqref{multi-task learning} is calculated internally
in the network using \eqref{best estimates C}.

We now assume that the discriminatory information $\bD_i$ is only available on part of the insurance policies
$i \in {\mathcal I} \subset \{1,\ldots, n\}$. This requires that we mask $\bD_i$ for the
policies $i \in {\mathcal I}^c=\{1,\ldots, n\} \setminus {\mathcal I}$ where no discriminatory
information is available.
As mask we set $\bD_i={\tt NA}$ for $i \in {\mathcal I}^c$. This will ignore the parts of the
following loss function for which the discriminatory information is not available.
We set for the
new optimization problem
\begin{eqnarray}\label{multi-task loss NA}
\widehat{\btheta} &=& \underset{\btheta}{\arg \min} \, \sum_{i=1}^n 
\Bigg[ \, \sum_{k=1}^K L_{\mu} \Big(Y_i, \,
\mu(\bX_i, \bd_k) \Big)\, \mathds{1}_{\{\bD_i = \bd_k\}}\\
&& \hspace{2cm} + ~ L_p \Big(\bD_i, \,
 \left(p_k(\bX_i)\right)_{1\le k \le K} \Big)\,\mathds{1}_{\{\bD_i \neq {\tt NA}\}} +L_{\mu} \Big(Y_i, \,
\mu(\bX_i) \Big) \Bigg],
\nonumber
\end{eqnarray}
for given loss functions $L_{\mu}$ and $L_p$. This network fitting problem \eqref{multi-task loss NA}
provides the interaction between the two FNNs, and it only considers the parts of the
loss function where the corresponding information is available.

\medskip

\begin{remark}\normalfont~
\begin{itemize}
\item
The multi-output network \eqref{multi-task learning} adds extra components
to predict the discriminatory information $\bD$ from the non-discriminatory $\bX$. Moreover, 
we directly determine the unawareness price $\mu(\bX)$, defined through \eqref{best estimates C},
and this is integrated into the last term of \eqref{multi-task loss NA}. This model can be fitted by
solving the optimization \eqref{multi-task loss NA}.
For the loss function $L_{\mu}$ we typically either use the square loss function or the deviance loss function
(within the exponential dispersion family), and for the loss function $L_p$ we are going to use
the multinomial cross-entropy loss. Intuitively, one should scale the loss functions $L_{\mu}$
and $L_{p}$ such that they live on a comparable scale. In our numerical examples, the results
have shown little sensitivity in such a scaling, therefore, we will just use the standard form
of the deviance losses below, i.e., without any additional scaling.
\item In our multi-task FNN we consider two parallel FNNs, and the interaction is only considered
by the joint parameter estimation in \eqref{multi-task loss NA}. We could also consider other network architectures
where, e.g., we learn a common representation $\bz^{(m:1)}(\bx)$ 
which serves to construct the 
readouts of $\mu(\bx, \bd_k)$ and $p_k(\bx)$, $1\le k \le K$. 
In our numerical experiments this latter approach was less competitive
in terms of predictive power
compared to the first one, but more work would be required
to come to a conclusive answer about the 'best' network architecture for this
multi-task learning problem.
\item The last component of the objective function
in \eqref{multi-task loss NA} compares the (internally) calculated
unawareness price $\mu(\bX)$ to the response $Y$. Alternatively, we could fit another 
regression model $\widetilde{\mu}(\bX)$ for the unawareness price. This can be
done because it does not involve any protected information $\bD$. A variant of 
\eqref{multi-task loss NA} then replaces the last objective function in \eqref{multi-task loss NA} 
as follows
\begin{eqnarray}\label{multi-task loss NA 2}
\widehat{\btheta} &=& \underset{\btheta}{\arg \min} \, \sum_{i=1}^n 
\Bigg[ \, \sum_{k=1}^K L_{\mu} \Big(Y_i, \,
\mu(\bX_i, \bd_k) \Big)\, \mathds{1}_{\{\bD_i = \bd_k\}}\\
&& \hspace{-.31cm} + ~ L_p \Big(\bD_i, \,
 \left(p_k(\bX_i)\right)_{1\le k \le K} \Big)\,\mathds{1}_{\{\bD_i \neq {\tt NA}\}} +L_{\widetilde{\mu}} \Big(
 \widetilde{\mu}(\bX_i), \,
\mu(\bX_i) \Big) \Bigg],
\nonumber
\end{eqnarray}
for given loss functions $L_{\mu}$, $L_{\widetilde{\mu}}$ and $L_p$. This approach can be 
useful in certain situations of over-fitting, but it requires a high-quality model for 
$\widetilde{\mu}$ in order to outperform the fitting procedure \eqref{multi-task loss NA}.
\item The multi-task FNN \eqref{multi-task learning} is solely used to 
model the discrimination-free insurance price through \eqref{discrimination-free price}.
There are many different notions and
definitions of fairness that may complement the discrimination-free insurance prices;
see, e.g., Grari et al.~\cite{GrariEtAl}. Multi-task learning
\eqref{multi-task learning}-\eqref{multi-task loss NA} can be extended by such
complementary fairness notions. This requires that the notion of the chosen fairness criteria
can be encoded into sensible scores that can be added
to the optimization \eqref{multi-task loss NA}, and depending on the quantities needed in these additional
scores, we may need to add corresponding outputs to the multi-task learning
\eqref{multi-task learning}. As a result the multi-output network will be regularized by the
corresponding scoring parts that account for the selected notions of fairness.
\item Alternatively to using certain scores for regularization, one can also set-up a 
generative-adversarial network (GAN), that tries to infer protected information from the
prices. A successful implementation of such a GAN architecture will force the pricing
network to design a pricing functional from which the gender cannot be inferred,
we also refer to Grari et al.~\cite{GrariEtAl}.
\end{itemize}
\end{remark}


\section{Synthetic health insurance example}
\label{Synthetic health insurance example}
We design a synthetic health insurance example that is similar to
Lindholm et al.~\cite{Lindholm}, but with a slightly more complicated underlying
regression function. Working with a synthetic example, that is, knowing
the true data generating model, has the advantage of being able to benchmark the estimated
models to the ground truth. 

\subsection{Data generation}
Let the discriminatory information $\bD \in \{\text{female}, \text{male}\}$ be the gender
of the policyholder. The non-discriminatory information $\bX = (X_1, X_2)^\top$
is assumed to have two components, with $X_1 \in \{15,16, \ldots, 80\}$ denoting the age of the policyholder
and $X_2\in \{\text{non-smoker}, \text{smoker}\}$ the smoking status of the policyholder.
There are different claim types:
claims that mainly affect females between ages 20 and 40
and males after age 60 (type 1), 
claims with a higher frequency for smokers and also for females (type 2), 
and general claims due to other disabilities (type 3). The logged expected frequencies
of these claim types are given by 
\begin{eqnarray*} 
  \log \lambda_1(\bX, \bD) &=& \alpha_0 + \alpha_1 \mathds{1}_{\{X_1 \in [20,40],\, \bD=\text{female}\}}+ \alpha_2 \mathds{1}_{\{X_1 \ge 60,\,\bD=\text{male}\}},\\
  \log \lambda_2(\bX, \bD) &=& \gamma_0 + \gamma_1X_1 + \gamma_2\mathds{1}_{\{X_2 = \text{smoker}\}} + \gamma_3 \mathds{1}_{\{\bD=\text{female}\}},\\
\log \lambda_3(\bX, \bD) &=& \delta_0 + \delta_1 X_1,
\end{eqnarray*}
with the following parameters: $(\alpha_0, \alpha_1, \alpha_2)= (-40, 38.5, 38.5)$, $(\gamma_0, \gamma_1,\gamma_2,\gamma_3)=( -2, 0.004, 0.1, 0.2)$, and $(\delta_0, \delta_1)= (-2, 0.01)$.  We set for the (true) total expected
claim frequency
\begin{equation*}
\lambda(\bX,\bD) = \lambda_1(\bX, \bD) + \lambda_2(\bX, \bD) + \lambda_3(\bX, \bD).
\end{equation*}
If we assume that the number of claims $Y$ of an insurance policyholder with covariates 
$(\bX, \bD)=(\bx, \bd)$ is Poisson distributed with expected value $\lambda(\bx,\bd)$,
then we obtain the (true) best-estimate price
\begin{equation}\label{true model}
\lambda(\bx,\bd) ~=~ \E \left[ Y \mid \bX=\bx, \bD=\bd \right],
\end{equation}
where, for simplicity, here we only focus on claim counts. Since in practice this true best-estimate price $\lambda(\bx,\bd)$ is not known,
we estimate it from the available data using a regression function denoted
by $\mu(\bx,\bd)$. We therefore first use a plain-vanilla FNN to model $\mu(\bx,\bd)$
based on the full input $(\bX, \bD)=(\bx, \bd)$, see \eqref{plain vanilla best estimate}.
In the next steps we study the multi-output
FNN \eqref{multi-output FNN} and the multi-task FNNs \eqref{multi-task learning}, which only use
the non-discriminatory information $\bX=\bx$ as a network input, while the available discriminatory
information $\bD$ is only used in the
loss functions for model fitting; see \eqref{multi-output fitting}, \eqref{multi-task loss NA}
and \eqref{multi-task loss NA 2}, respectively. Table \ref{table of models}
illustrates all models that we are going to consider, the labeling (a)-(f) will be kept
throughout this example.

\begin{table}[htb]
\begin{center}
{\small
\begin{tabular}{|c|l|}
\hline 
label & regression model \\
\hline
  (a) &true model with regression function $\lambda(\bx, \bd)$ given by \eqref{true model}\\
  (b) &plain-vanilla FNN regression function $\mu(\bx, \bd)$ given by \eqref{plain vanilla best estimate} \\
  (c) &multi-output FNN regression function $\mu(\bx, \bd_k)$ given by \eqref{multi-output FNN} \\
 (d)  &multi-task FNN ($Y$) regression function given by \eqref{multi-task learning}
 and \eqref{multi-task loss NA} \\
 (e)  &multi-task FNN ($\widehat{\mu}$) regression function given by \eqref{multi-task learning} and \eqref{multi-task loss NA 2}
  \\
\hline
(f)& regression tree boosting (best-estimate benchmark) \\
\hline
\end{tabular}}
\end{center}
\caption{Table of considered models fitting approaches.}
\label{table of models}
\end{table}

To fit these FNNs we first need to generate i.i.d.~data $(Y_i,\bX_i,\bD_i)_{1\le i \le n}$.
We select a portfolio of sample size $n=100,000$ as follows. The age variable $X_{1}$ is assumed to be
independent of the smoking habits $X_{2}$ and the gender $\bD$, and we choose
the age distribution as given in Figure 4 of Lindholm et al.~\cite{Lindholm}. 
Moreover, we choose $\P[\bD={\rm female}]=0.45$, 
$\P[X_2={\rm smoker}]=0.3$ and $\P[\bD={\rm female}\mid X_2={\rm smoker}]=0.8$. This
fully specifies the distribution of the covariates $(\bX,\bD)$, making smoking more common
among females compared to males in this population. We simulate $n=100,000$ independent insurance policies from this covariate distribution. This provides us with an empirical proportion of females in the simulated
data of 0.4505, fairly close to the true ratio of 0.45. Later, we use
this proportion for obtaining the pricing distribution, i.e., we set $\P^*[\bD={\rm female}]=0.4505$;
as explained in Lindholm et al.~\cite{Lindholm}, this is a canonical choice
for $\P^*$. Finally, we simulate independent observations
$Y_i|_{\{\bX_i,\bD_i\}}\sim {\rm Poi}(\lambda(\bX_i,\bD_i))$, $1\le i \le n$, giving the (pseudo-)sample $(Y_i,\bX_i,\bD_i)_{1\le i \le n}$, representing the portfolio of  policies. In the next section
we assume the availability of full covariate information for the protected characteristics $\bD_i$, whereas in Section \ref{Partial information about protected characteristics}
we will assume only partial access to such information.

\subsection{Full availability of discriminatory information}

\subsubsection{Plain-vanilla feed-forward neural network}
\label{Plain-vanilla feed-forward neural network}
We start with the plain-vanilla FNN \eqref{plain vanilla best estimate}, which takes as input the covariates
$(\bX,\bD) \in \R^{q_0}$ with $q_0=3$; we use dummy coding
for both the gender variable $\bD$ and the smoking habits $X_2$. We choose a network of depth $m=3$ with
$(q_1,q_2,q_3)=(20,15,10)$ hidden neurons in the three hidden layers, the ReLU activation function, and the
log-link $g(\cdot)=\log(\cdot)$, which is the canonical link of the Poisson regression model.
This network has a parameter $\btheta=(\bbeta, \bw)$ of dimension 566.
To implement this FNN we use the library {\tt Keras} \cite{keras} within
the statistical computing software {\sf R} \cite{R}.

We fit this FNN to the simulated data $(Y_i,\bX_i,\bD_i)_{1\le i \le n}$; assuming for now access to the full policyholder information, including $\bD$. 
We use the Poisson deviance loss function
for $L$ in \eqref{plain vanilla fitting}, the {\tt nadam} version of stochastic
gradient descent, a batch size of 50 policies, and we explore early stopping
based on a 80/20 training-validation split. This is similar to
Section 7.3.2 in W\"uthrich--Merz \cite{WM2021}; for more details
we refer to that source. Since network fitting involves several elements of randomness,
see Remarks 7.7 in W\"uthrich--Merz \cite{WM2021}, we average over 10 different
FNN calibrations, resulting in the nagging predictor of Richman--W\"uthrich \cite{RichmanW2}.

\begin{figure}[htb!] 
\begin{center}
\begin{minipage}{0.495\textwidth}
\begin{center}
\includegraphics[width=\linewidth]{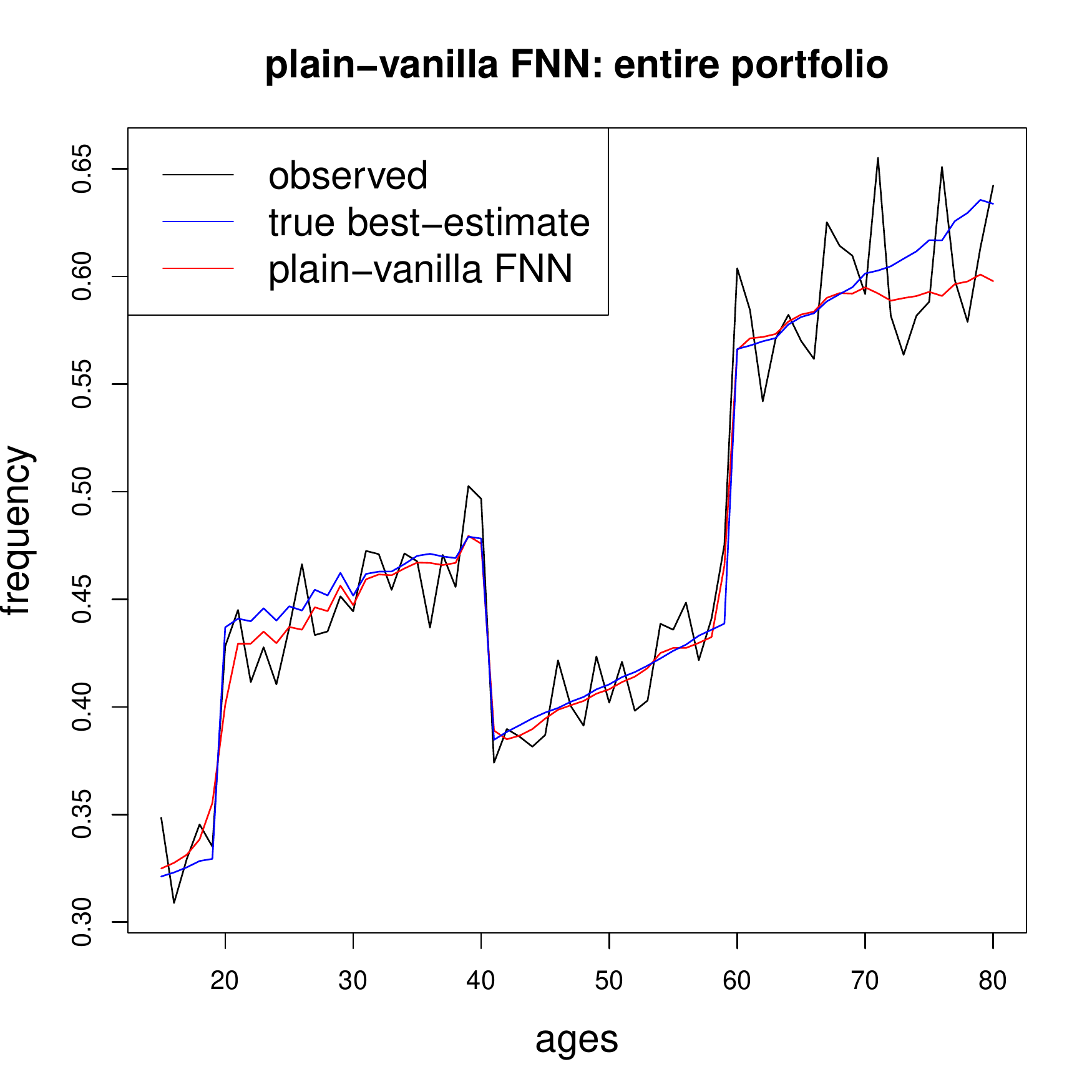}
\end{center}
\end{minipage}
\end{center}
\caption{Best-estimate price $\mu(\bx,\bd)$ 
of the fitted plain-vanilla FNN \eqref{plain vanilla best estimate} (red) compared to the
true best-estimate price $\lambda(\bx,\bd)$ (blue) as a function of the age 
variable $x_1$ and averaged over
smoking habits $x_2$ and gender $\bd$; these results
are based on the complete knowledge of discriminatory information. Raw observations are represented by the black line.}
\label{Figure: plain-vanilla best-estimate}
\end{figure}

The results are given in Figure 
\ref{Figure: plain-vanilla best-estimate}. The blue line shows the true
best-estimate price $\lambda(\bx, \bd)$ as a function of the age variable $15 \le x_1 \le 80$, averaged over the smoking habits $x_2$ and the gender variable $\bd$ w.r.t.~the empirical population density. The black color shows the corresponding observations $Y_i$ and the 
red color the plain-vanilla FNN fitted best-estimate price $\mu(\bx, \bd)$ using the
full inputs $(\bX_i, \bD_i)$. Overall, Figure \ref{Figure: plain-vanilla best-estimate} suggests a rather accurate fit; only at the age boundaries
there are some differences which are caused by the noise in the observations $Y_i$.

Since we know the true regression function $\lambda(\bx,\bd)$, we can
explicitly quantify the accuracy of the estimated FNN regression function $\mu(\bx,\bd)$.
That is, we do not need to validate the estimated model on a
test data sample, but we can directly compare it to the true model.
We use the Kullback--Leibler (KL) divergence to compare the estimated
model to the true one.
In the case of the Poisson model the KL divergence for given covariate values $(\bx, \bd)$ is given by
\begin{eqnarray}\nonumber
D_{\rm KL}\big(\lambda(\bx,\bd) \big|\big| \mu(\bx,\bd) \big)
&=& \sum_{y \in \N_0} e^{-\lambda(\bx,\bd)}\frac{\lambda(\bx,\bd)^y}{y!}
\log \left( \frac{e^{-\lambda(\bx,\bd)}\frac{\lambda(\bx,\bd)^y}{y!}}{e^{-\mu(\bx,\bd)}\frac{\mu(\bx,\bd)^y}{y!}}\right)
\\&=& \mu(\bx,\bd) -\lambda(\bx,\bd) 
- \lambda(\bx,\bd) \log\left(\frac{\mu(\bx,\bd)}{\lambda(\bx,\bd)}\right).
\label{KL divergence individual}
\end{eqnarray}
We average the KL divergence of a single instance $(\bx, \bd)$ over the empirical population 
distribution, which gives us the KL divergence from the estimated model to the true model
on our portfolio
\begin{equation}\label{KL divergence}
D_{\rm KL}\left(\lambda || \mu \right)
:= \frac{1}{n} \sum_{i=1}^n D_{\rm KL}\big(\lambda(\bx_i,\bd_i) \big|\big| \mu(\bx_i,\bd_i) \big).
\end{equation}
This gives us a measure of model accuracy for the different estimated models;
the KL divergence is zero if and only if the estimated model is identical to
the true model on the selected portfolio; see Section 2.3 in W\"uthrich--Merz \cite{WM2021}.

\begin{table}[htb]
\begin{center}
{\small
\begin{tabular}{|l||c|}
\hline 
 & KL divergence \eqref{KL divergence}  \\
 & to $\lambda(\bx,\bd)$\\
\hline\hline
(b) plain-vanilla FNN: full data  & 0.2204 \\
(c) multi-output FNN: full data  & 0.2567 \\
(d) multi-task FNN ($Y$): full data  & 0.2823 \\
(e) multi-task FNN ($\widehat{\mu}$): full data  & 0.3070 \\
\hline
(f) regression tree boosting: full data & 0.5170 \\
\hline

\end{tabular}}
\end{center}
\caption{Model accuracy of the fitted FNNs and a regression tree boosting model serving as
  a benchmark; the KL divergences are stated in $10^{-3}$.}
\label{model accuracy}
\end{table}

Row (b) of Table \ref{model accuracy} shows the KL divergence \eqref{KL divergence}
of the fitted plain-vanilla FNN best-estimate price $\mu(\bx,\bd)$ to the true 
best-estimate price $\lambda(\bx,\bd)$. The resulting KL divergence is $0.2204\cdot 10^{-3}$, which is
much smaller than a comparable regression tree boosting model that results in a KL
divergence of $0.5170\cdot 10^{-3}$, shown in row (f) of Table \ref{model accuracy}.

This fitted FNN can now
be used for discrimination-free insurance pricing according to 
formula \eqref{discrimination-free price} for the given selected measure $\P^*$. However,
we cannot calculate the unawareness price $\mu(\bx)$ from \eqref{unawareness price} because this
requires the knowledge of the probabilities $(p_k(\bx))_{1\le k \le K}$, see \eqref{tower property}. Alternatively, we could directly fit a plain-vanilla FNN for estimating the unawareness price, a route we do not pursue here. We come back to this topic when discussing the multi-task FNN.

\subsubsection{Multi-output feed-forward neural network}
\label{Multi-output feed-forward neural network}
\begin{figure}[htb!] 
\begin{center}
\begin{minipage}{0.495\textwidth}
\begin{center}
\includegraphics[width=\linewidth]{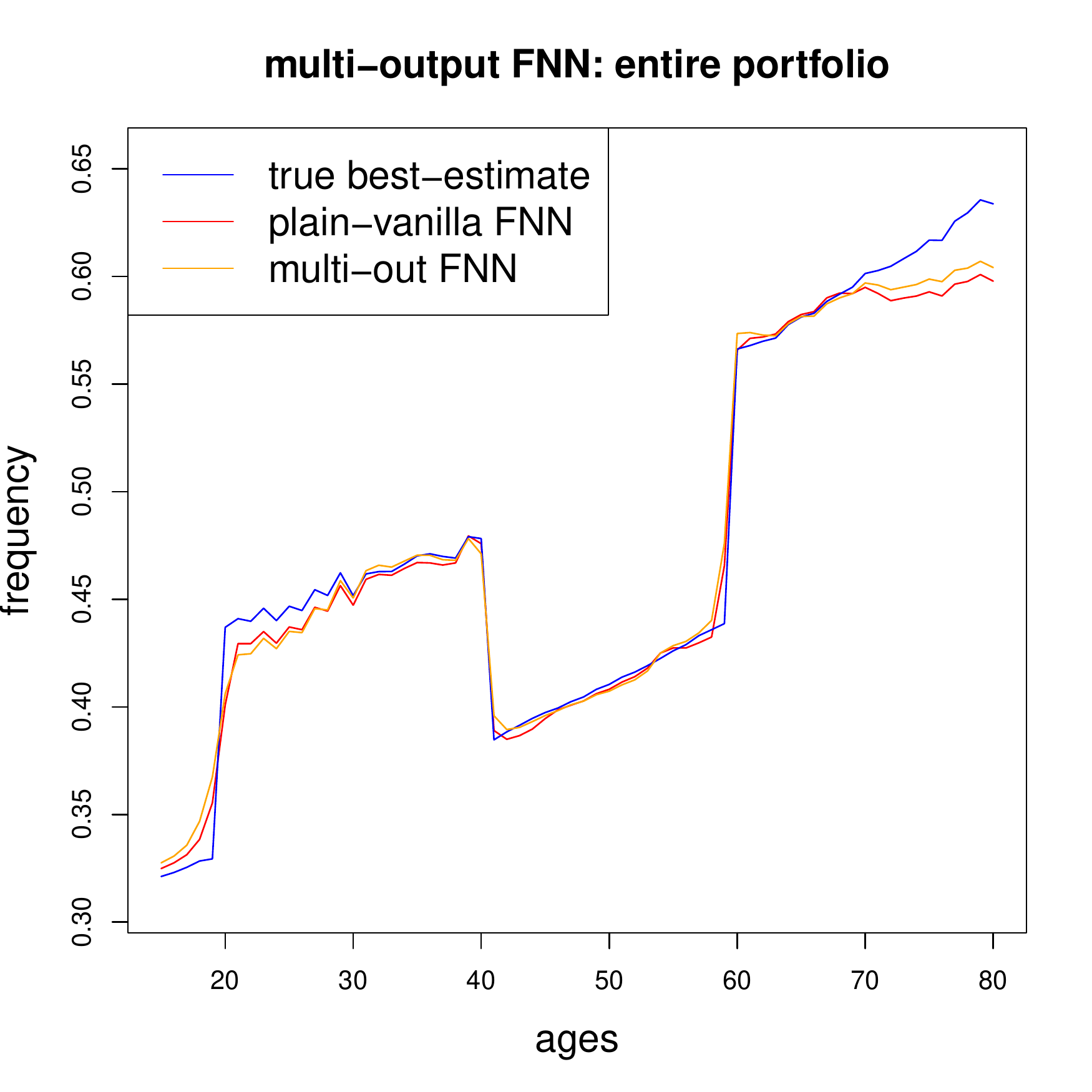}
\end{center}
\end{minipage}
\begin{minipage}{0.495\textwidth}
\begin{center}
\includegraphics[width=\linewidth]{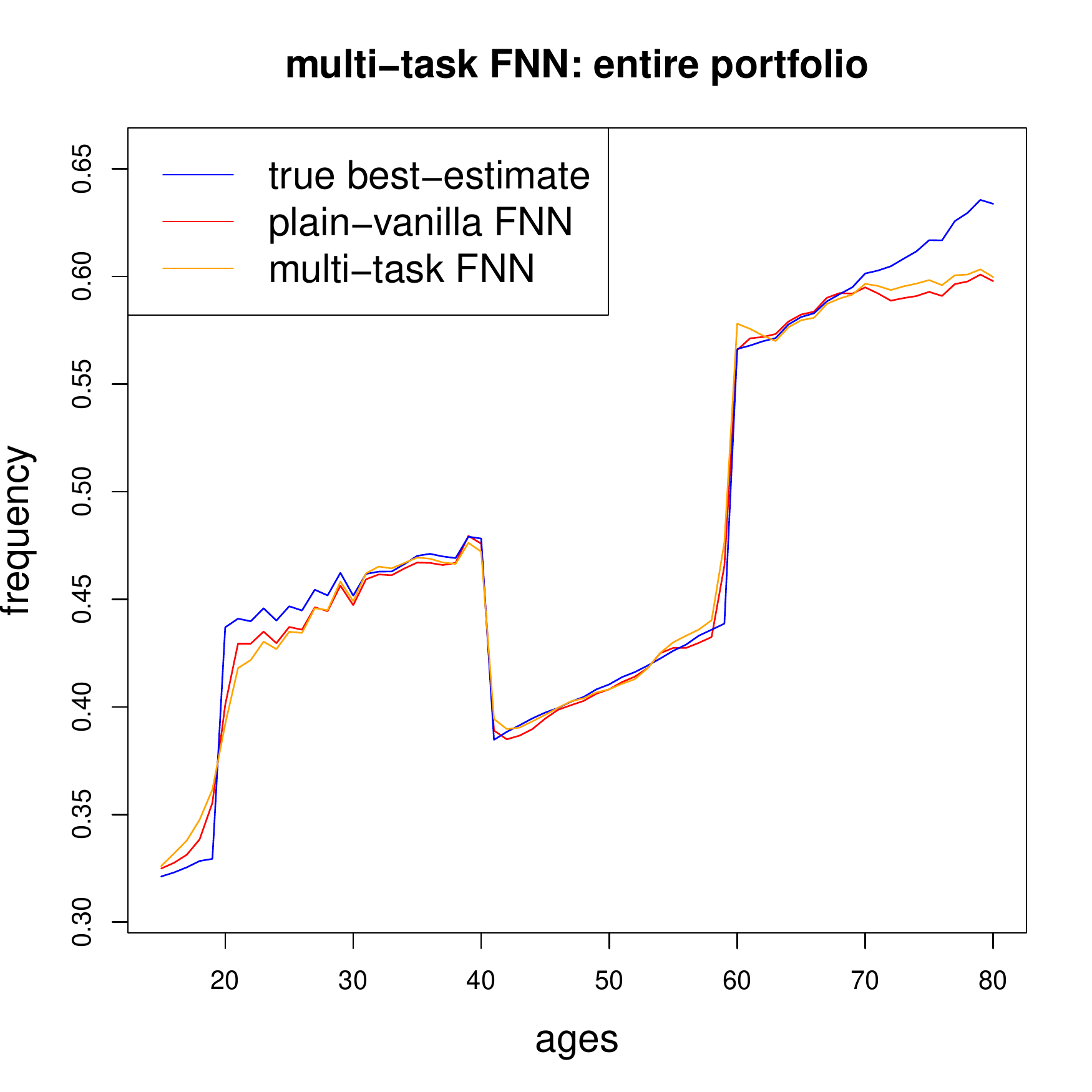}
\end{center}
\end{minipage}
\end{center}
\caption{Best-estimate prices $(\mu(\bx,\bd_k))_{1\le k \le K}$ 
of the fitted multi-output FNN \eqref{multi-output FNN} (lhs) and
the fitted multi-task FNN \eqref{multi-task learning}
with loss \eqref{multi-task loss NA} (rhs) (orange); compared to the estimated prices of the plain-vanilla FNN (red) and the true best-estimates (blue). The estimates are based on complete knowledge of discriminatory information.}
\label{Figure: multi-task best-estimate}
\end{figure}

Next, we fit the multi-output FNN \eqref{multi-output FNN} to the same data,
only using the non-discriminatory covariates $\bx$ as input to the network. 
This reduces the input dimension to $q_0=2$, but on the other
hand we have two outputs $\mu(\bx, \bd={\rm female})$ and
$\mu(\bx, \bd={\rm male})$ in the multi-output FNN. The former reduces the dimension of the network
parameter and the latter increases the dimension 
of the network parameter,  resulting in a network
parameter $\btheta=(\bbeta_{\rm female},\bbeta_{\rm male}, \bw)$ of dimension 557.
We fit this multi-output FNN using exactly the same fitting strategy as above.
The results are presented in orange color in Figure \ref{Figure: multi-task best-estimate} (lhs),
and they are compared to the plain-vanilla FNN best-estimates in red color and the
true best-estimates in blue color.
We conclude that the two networks (orange and red) provide rather similar results.

Table \ref{model accuracy} shows a slightly higher accuracy of the plain-vanilla FNN (row (b))
compared to the
multi-output FNN (row (c)), with respective KL divergence values of  $0.2204\cdot 10^{-3}$ vs.~$0.2567\cdot 10^{-3}$. In general, these numbers are
quite small and the models are rather accurate, which is in support of both FNN models.

\subsubsection{Multi-task feed-forward neural network}
\label{Model accuracy}
We now fit the multi-task FNN \eqref{multi-task learning} to the data,
still assuming full knowledge of the discriminatory information $\bD_i$.
The multi-task FNN additionally
predicts the discriminatory covariates $\bD_i$ which are then (internally) used to
calculate the unawareness price $\mu(\bx)$, see \eqref{best estimates C}-\eqref{multi-task learning}.
That is, in contrast to the plain-vanilla
FNN and the multi-output FNN, this is the only one of the three approaches that allows
us to directly calculate the unawareness prices within the {\it same} model 
as the best-estimate prices. 
We start with objective function \eqref{multi-task loss NA} considering the
response $Y$ for assessing the unawareness price $\mu(\bx)$.

We use exactly the same fitting strategy
as in the previous two modeling approaches. Figure \ref{Figure: multi-task best-estimate} (rhs)
shows the resulting best-estimate prices (in orange color) compared to the ones
of the plain-vanilla FNN (in red color). Row (d) of Table \ref{model accuracy} provides a
resulting KL divergence to the true model of $0.2823\cdot 10^{-3}$. This is slightly higher
than in the other two approaches, but still gives a very competitive result.
The full advantage of the multi-task FNN approach will become clear once we start
working with incomplete discriminatory information in Section 
\ref{Partial information about protected characteristics} below.

Finally, we present the fitting results of the multi-task FNN \eqref{multi-task learning}
when using objective function \eqref{multi-task loss NA 2}. For this we first fit a plain-vanilla
FNN $\widetilde{\mu}(\bx)$ to the unawareness price (only considering $\bx$).
This is done completely analogously
to Section \ref{Plain-vanilla feed-forward neural network} except that we drop the discriminatory
information from the input. We then use this fitted FNN $\widetilde{\mu}(\bx)$ in objective function 
\eqref{multi-task loss NA 2}, and we use the KL divergence \eqref{KL divergence individual}
for $L_{\widetilde{\mu}}$
to measure the divergence from $\mu(\bx)$ to $\widetilde{\mu}(\bx)$.
The results are presented on row (e) of Table \ref{model accuracy}. We observe that this
is the least accurate of all FNN models, the main issue probably being that the regression function
$\widetilde{\mu}(\bx)$ is not sufficiently accurate, and we should rather directly
compare the unawareness price $\mu(\bx)$
 to the observations $Y$ as done in \eqref{multi-task loss NA}.
For this reason, we will not further pursue this approach below.

\subsubsection{Discrimination-free insurance pricing}
\begin{figure}[htb!] 
\begin{center}
\begin{minipage}{0.495\textwidth}
\begin{center}
\includegraphics[width=\linewidth]{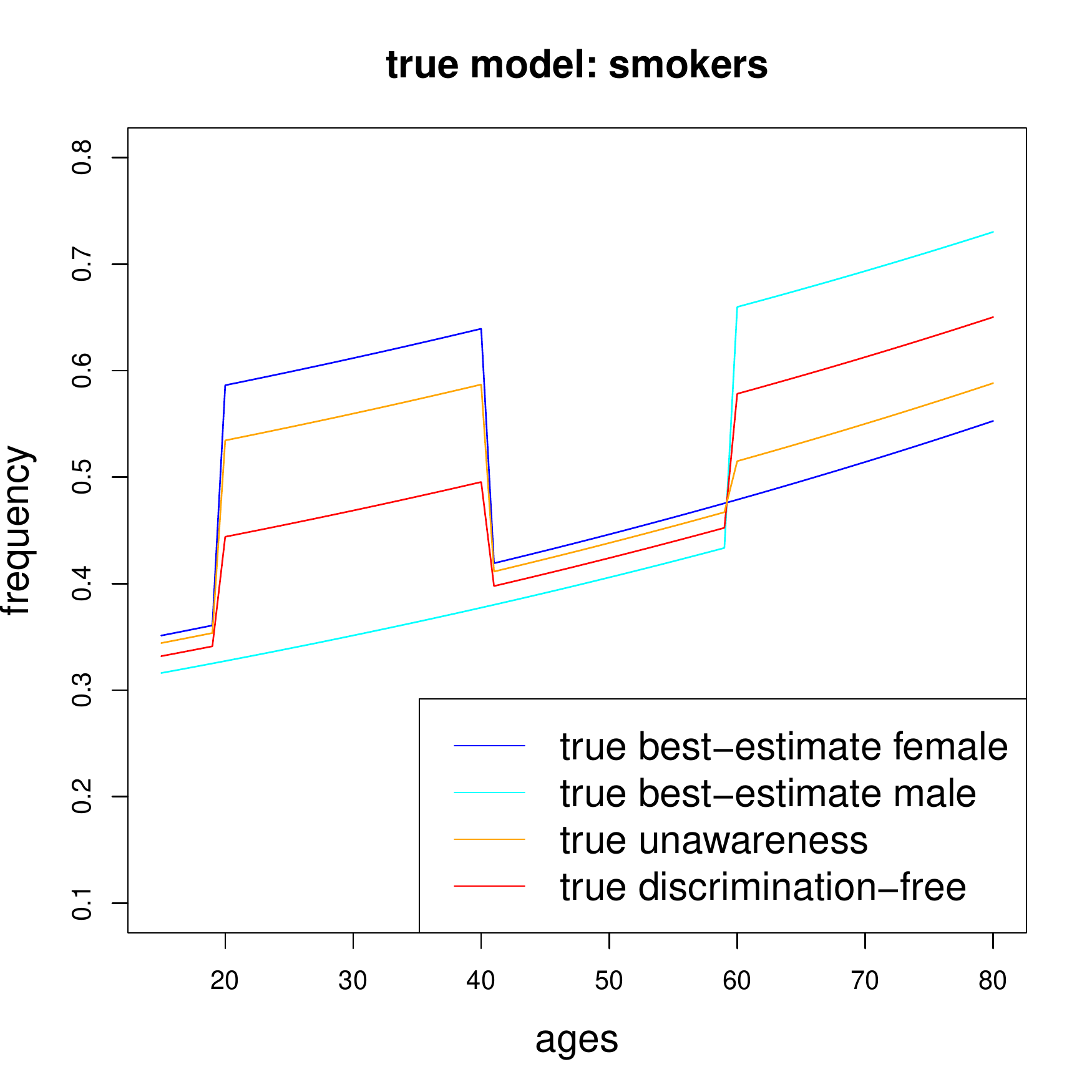}
\end{center}
\end{minipage}
\begin{minipage}{0.495\textwidth}
\begin{center}
\includegraphics[width=\linewidth]{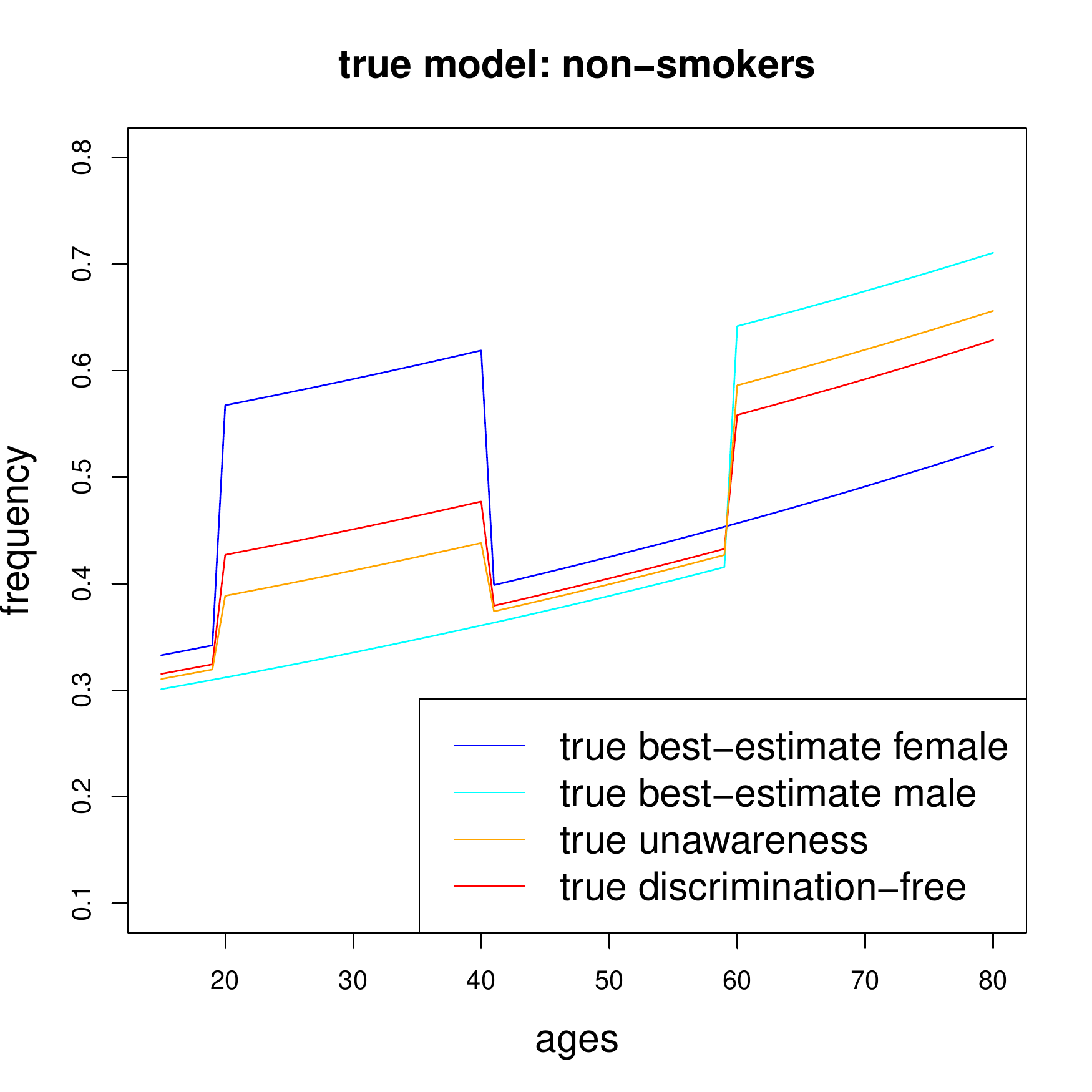}
\end{center}
\end{minipage}
\end{center}
\caption{True model: best-estimate prices 
  $\lambda(\bx,\bd)$, unawareness prices $\lambda(\bx)$ and discrimination-free insurance prices
  $\lambda^*(\bx)$ with
(lhs) smokers, and (rhs) non-smokers.}
\label{Figure: true DFIP}
\end{figure}

Having fitted the three FNNs, we can calculate the
discrimination-free insurance prices $\mu^*(\bx)$ by \eqref{DFIP discrete}, using the empirical gender distribution
$\P^*[\bD={\rm female}]=0.4505$ as the pricing distribution. 
We start by considering the true model $\lambda(\bx,\bd)$. We calculate
the discrimination-free insurance prices $\lambda^*(\bx)$ and the unawareness prices $\lambda(\bx)$
in the true model. This can be done because all necessary information is available.
The corresponding graphs are shown in Figure \ref{Figure: true DFIP}, and they  will serve
as a benchmark for the estimated FNNs. It is seen that the unawareness prices $\lambda(\bx)$ closely follow
the best-estimate prices $\lambda(\bx,\bd)$ of females for smokers, and the best-estimate prices of males for
non-smokers. This reflects the fact that, in our example, smoking habits are rather informative for predicting the gender.
The discrimination-free insurance prices $\lambda^*(\bx)$ exactly correct for this inference potential: as seen from Figure \ref{Figure: true DFIP}, while smokers of either gender have higher predicted claims frequencies than non-smokers, discrimination-free insurance prices lie between the best-estimate prices for males and females, following the same pattern regardless of smoking status.

\begin{table}[htb]
\begin{center}
{\small
\begin{tabular}{|l||c|}
\hline 
 & KL divergence \eqref{KL divergence}  \\
 & to $\lambda(\bx,\bd)$\\
\hline\hline
  (a0) true unawareness price $\lambda(\bx)$ & 6.3174 \\
  (d0) multi-task FNN ($Y$) unawareness price $\mu(\bx)$ & 6.4932 \\
  \hline
  (a1) true discrimination-free price $\lambda^*(\bx)$ & 7.8857 \\
  (b1) plain-vanilla FNN discrimination-free price $\mu^*(\bx)$ & 8.3222 \\
  (c1) multi-output FNN discrimination-free price $\mu^*(\bx)$ & 8.2669 \\
 (d1)  multi-task FNN ($Y$) discrimination-free price $\mu^*(\bx)$ & 8.2915 \\
  
\hline

\end{tabular}}
\end{center}
\caption{Model accuracy of the unawareness prices $\mu(\bx)$ and the discrimination-free 
insurance prices
  $\mu^*(\bx)$;
  the KL divergences to the best-estimate $\lambda(\bx,\bd)$
  are stated in $10^{-3}$. The figures are based on the
  full knowledge of discriminatory information.}
\label{model accuracy 2}
\end{table}

Table \ref{model accuracy 2} presents the corresponding numerical results. The unawareness price 
$\lambda(\bx)$ in the true model has a KL divergence to the best-estimate price $\lambda(\bx,\bd)$ of
$6.3174 \cdot 10^{-3}$, see row (a0). That is, we sacrifice quite some predictive accuracy by ignoring the
protected information $\bD_i$ in the unawareness price $\lambda(\bx)$.
This approach still internally infers the gender
from the non-discriminatory information, see \eqref{tower property}. Breaking this link further
decreases the predictive accuracy, resulting in a KL divergence from the discrimination-free 
insurance price
$\lambda^*(\bx)$ to the best-estimate price $\lambda(\bx,\bd)$
of $7.8857 \cdot 10^{-3}$, see Table \ref{model accuracy 2}, row (a1).

Furthermore, Table \ref{model accuracy 2} provides all KL divergences to the
true best-estimate price $\lambda(\bx,\bd)$  that can be calculated
from the three fitted FNNs: row (b1) considers the discrimination-free insurance price $\mu^*(\bx)$ in the plain-vanilla FNN \eqref{plain-vanilla FNN}, row (c1) the discrimination-free insurance price $\mu^*(\bx)$  in the
multi-output FNN \eqref{multi-output FNN} and 
rows (d0) and (d1) the unawareness price $\mu(\bx)$ and the discrimination-free insurance price $\mu^*(\bx)$  in 
the multi-task FNN \eqref{multi-task learning} using objective function \eqref{multi-task loss NA}.
The last FNN is the only one that directly provides the unawareness price $\mu(\bx)$. 
The accuracy of the resulting discrimination-free insurance prices $\mu^*(\bx)$ is rather similar between the three network approaches (KL divergences on rows (b1)-(d1) of Table \ref{model accuracy 2}). Clearly we sacrifice 
quite some predictive power by not being allowed to use the protected
information $\bD_i$, such that the KL divergences increase from $0.25 \cdot 10^{-3}$ in
Table \ref{model accuracy} to roughly
$8 \cdot 10^{-3}$ in
Table \ref{model accuracy 2}.

\begin{table}[htb]
\begin{center}
{\small
\begin{tabular}{|l||c|}
\hline 
 & KL divergence \eqref{KL divergence}  \\
 & to $\lambda^*(\bx)$\\
\hline\hline
  (b2) plain-vanilla FNN discrimination-free price $\mu^*(\bx)$ & 0.1748 \\
  (c2) multi-output FNN discrimination-free price $\mu^*(\bx)$ & 0.1885 \\
 (d2)  multi-task FNN discrimination-free price $\mu^*(\bx)$ & 0.2323 \\
  
\hline

\end{tabular}}
\end{center}
\caption{Model accuracy of the discrimination-free insurance prices
  $\mu^*(\bx)$ relative to the true discrimination-free insurance price $\lambda^*(\bx)$;
  the KL divergences are stated in $10^{-3}$ and these figures are based on the
  full knowledge of discriminatory information.}
\label{model accuracy 3}
\end{table}

Table \ref{model accuracy 3} compares in KL divergence the discrimination-free 
insurance prices $\mu^*(\bx)$
of the three fitted FNNs to the true discrimination-free insurance price $\lambda^*(\bx)$.
The plain-vanilla FNN provides slightly more accurate results compared to the 
multi-output and the multi-task FNNs.
Note that the different rankings in Tables \ref{model accuracy 2}
and \ref{model accuracy 3} may be caused by the randomness in the data
and by potentially different over- or under-fitting to the data. A
verification of the explicit reasons for 
these different rankings is difficult, and different samples may also change this order.

\begin{figure}[htb!] 
\begin{center}
\begin{minipage}{0.495\textwidth}
\begin{center}
\includegraphics[width=\linewidth]{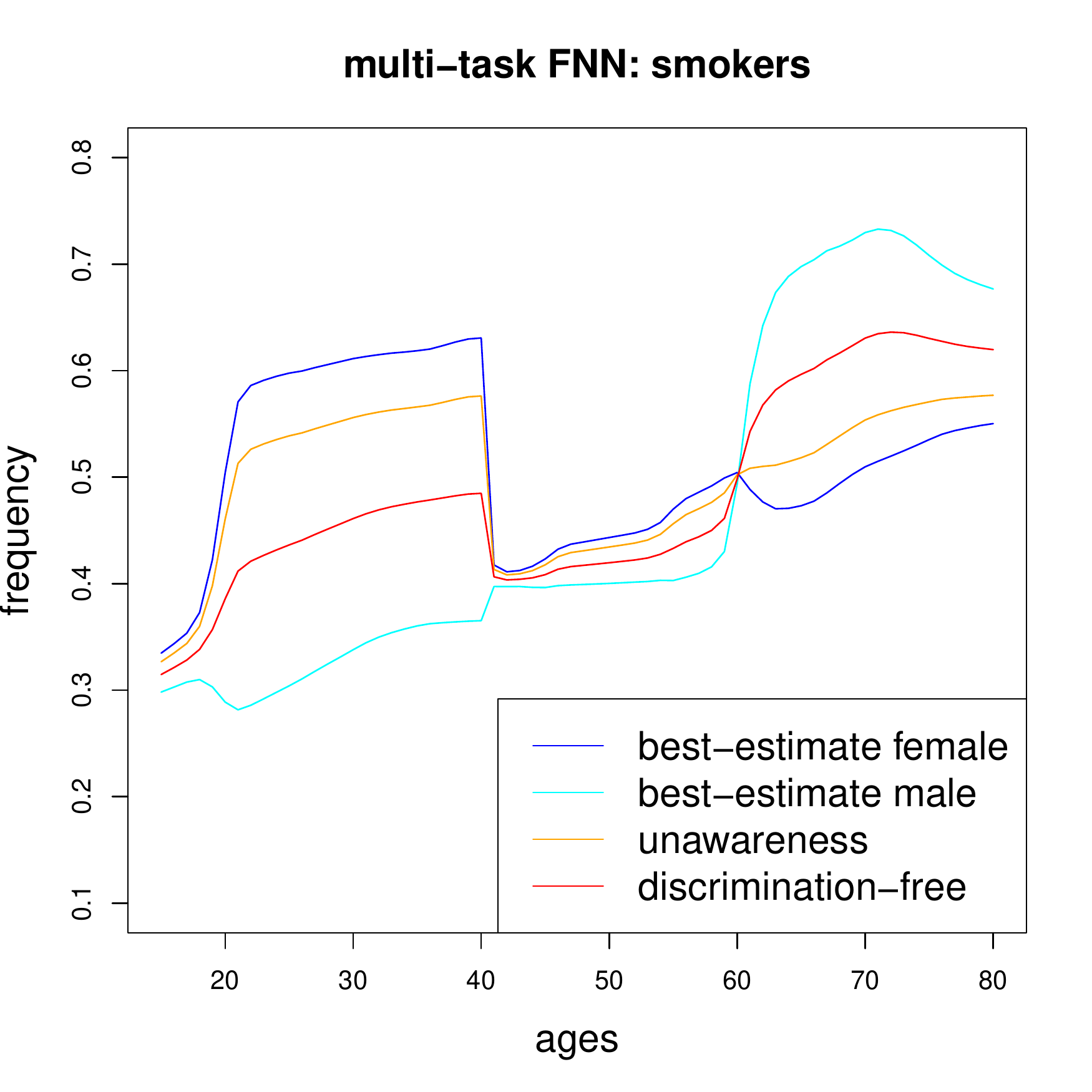}
\end{center}
\end{minipage}
\begin{minipage}{0.495\textwidth}
\begin{center}
\includegraphics[width=\linewidth]{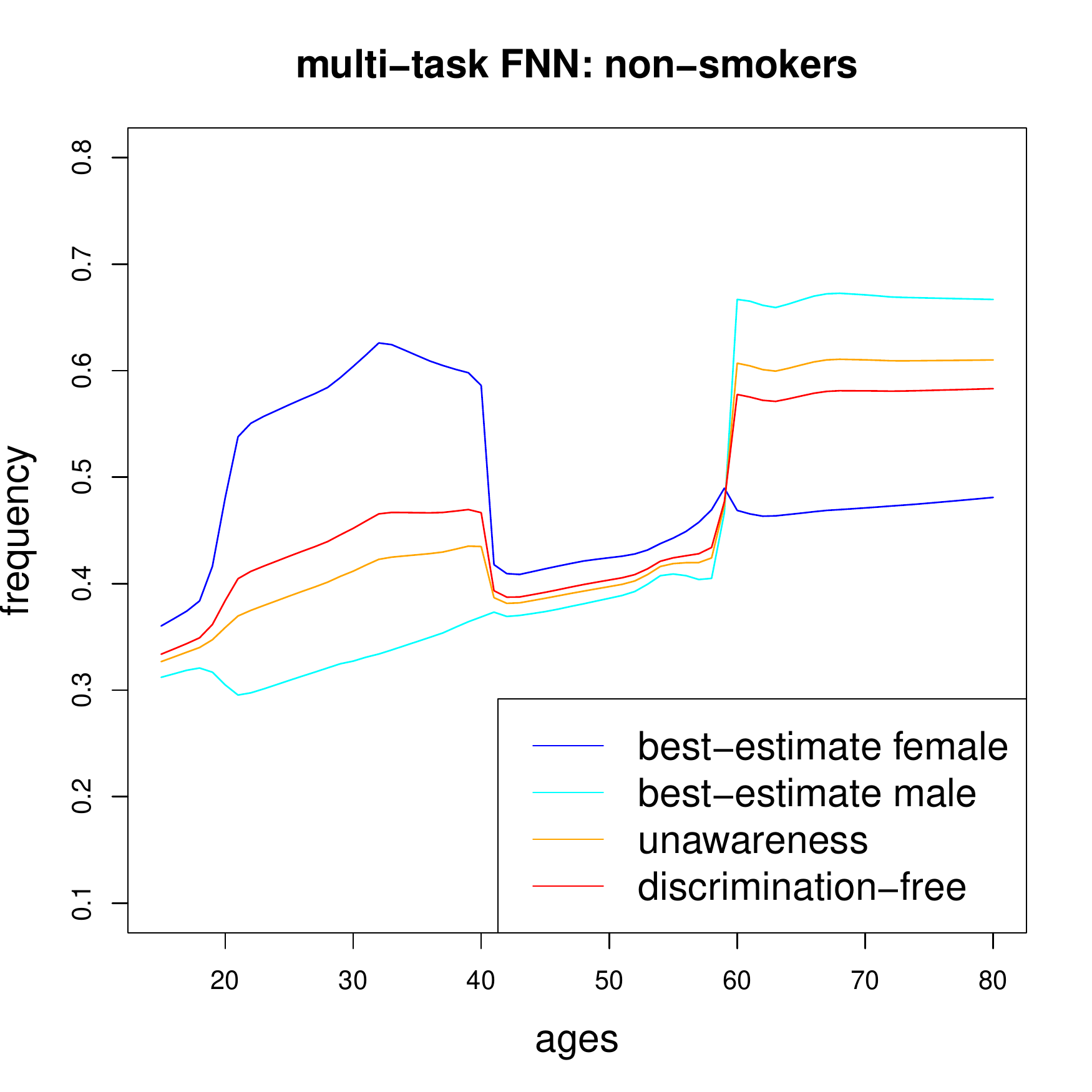}
\end{center}
\end{minipage}
\end{center}
\caption{Multi-task FNN: best-estimate prices 
  $\mu(\bx,\bd)$, unawareness prices $\mu(\bx)$ and discrimination-free insurance prices
  $\mu^*(\bx)$ with
(lhs) smokers, and (rhs) non-smokers.}
\label{Figure: multi-task DFIP}
\end{figure}

Finally, Figure \ref{Figure: multi-task DFIP} illustrates the resulting prices from the multi-task FNN.
They should be compared to the true ones in Figure \ref{Figure: true DFIP}. The interpretation is the same
in the two figures. Comparing the two plots we can also clearly see the impact of model uncertainty
in Figure \ref{Figure: multi-task DFIP}, which can only be mitigated by having larger sample sizes.

\subsubsection{Quantification of direct and indirect discrimination}
The results of Tables \ref{model accuracy}-\ref{model accuracy 3} give motivation 
to quantify the potential for direct and indirect discrimination. In contrast to these former results,
we now compare the unawareness price $\mu(\bx)$ and the discrimination-free 
insurance price $\mu^*(\bx)$ to the best-estimate price $\mu(\bx,\bd)$ {\it within} a given
model. 
The step from the best-estimate price $\mu(\bx,\bd)$ to the unawareness price 
$\mu(\bx)$ accounts for direct discrimination by quantifying 
the effect of simply being blind w.r.t.~the protected
information $\bD$.  Going from the unawareness
price $\mu(\bx)$ to the discrimination-free insurance price $\mu^*(\bx)$ accounts for indirect discrimination.
However, these steps are more
subtle for several reasons. First, the KL divergence does not satisfy the triangle inequality and, hence, divergences cannot simply be decomposed along a certain path.
Second, in some models we cannot
simultaneously calculate the best-estimate, the unawareness and the discrimination-free
insurance prices, but need to use different models to estimate these 
quantities. This applies, e.g., to the multi-output
FNN where we do not receive the unawareness price within that network model, 
but we have to explore another (separate)
model to estimate this unawareness price. This critical point does not apply to the multi-task
FNN, where we consistently calculate all terms within the same model.
Third, interpreting the potential
for indirect discrimination more broadly, there are two ingredients, namely, the inference
part $\P[\bD=\bd| \bX]$ and the best-estimate prices $\mu(\bx,\bd)$. Only if both of them are sufficiently
imbalanced, indirect discrimination becomes relevant (and visible). If we think of different insurance companies
selling the same product (with the same underwriting standards and the same claim costs), 
these companies should use the same best-estimate prices $\mu(\bx,\bd)$.
Indirect discrimination will typically differ between these companies, because
they will generally have
 different portfolio distributions $\P(\bX,\bD)$, resulting
in different inference potentials. Thus, the statistical dependence between $\bX$ and $\bD$
is company-specific, and so is the amount of indirect discrimination.

\begin{table}[htb]
\begin{center}
{\small
\begin{tabular}{|l||cc|}
\hline 
& \multicolumn{2}{|c|}{KL divergences to the best-estimates:}\\
& unawareness & discrimination-free\\
\hline\hline
 (a) true model & 6.3174  & 7.8857 \\
& 100\% & 125\%\\
 \hline
 (d)  multi-task FNN & 6.7980  & 8.5339 \\
& 100\% & 126\%  \\
\hline

\end{tabular}}
\end{center}
\caption{Losses of model accuracy using the unawareness price and the 
discrimination-free insurance price instead of the best-estimate price;
  the KL divergences are stated in $10^{-3}$.}
\label{discrimination potential}
\end{table}

Table \ref{discrimination potential} reflects the loss of model accuracy
if we deviate from the best-estimate price. These losses can be interpreted as the 
quantification of direct and indirect discrimination. If we normalize the 
KL divergence of the unawareness price to
100\%, then the discrimination-free insurance price adds another 25\% to
the loss of model accuracy compared to the unawareness price. 
Thus, direct discrimination is clearly the dominant
term, here. Nonetheless, we not that the numbers in Table \ref{discrimination potential}
reflect portfolio considerations, and the impact of indirect discrimination on particular
sub-populations or individual policies may be more substantial.

\subsection{Partial availability of discriminatory information}
\label{Partial information about protected characteristics}
\subsubsection{Missing completely at random}
So far, all numerical results have been based on
full knowledge of the data $(Y_i,\bX_i,\bD_i)_{1\le i \le n}$.
Next, we turn our attention to the problem of having incomplete discriminatory information,
and analyze how well we can fit our FNNs under this partial information setting. We therefore randomly remove
$\bD_i$ from the information set. This is done by independently (across the entire portfolio)
setting $\bD_i={\tt NA}$ with increasing (drop-out) probabilities of $10\%, 20\%, \ldots, 90\%$. That is, in the
last case only roughly 10\% of the discriminatory labels $\bD_i$ are available, and in the first
case roughly 90\% of all discriminatory labels are available.
Working with these drop-outs also changes the empirical female ratio that can
be calculated on the policies with full information. We state these in Table
\ref{female ratio} (row `empirical $\P(\bd)$')
as we use them for the pricing measure $\P^*(\bd)$.

\begin{table}[htb]
\begin{center}
{\scriptsize
\begin{tabular}{|l|cccccccccc|}
\hline 
drop-out & 90\%&80\%&70\%&60\%&50\%&40\%&30\%&20\%&10\%& 0\%\\\hline
available $\bD_i$ & 10\%&20\%&30\%&40\%&50\%&60\%&70\%&80\%&90\%&100\%\\\hline
empirical ${\P}(\bd)$ & 44.8\%& 44.5\%& 44.3\%& 44.5\%& 44.6\%& 45.5\%& 45.4\%& 45.2\%& 45.1\%&  45.1\%\\
multi-task $\widehat{\P}(\bd)$ & 45.8\%& 44.8\%& 44.3\%& 44.5\%& 44.7\%& 45.5\%& 45.4\%& 45.2\%& 45.0\%& 45.1\%\\
  \hline

\end{tabular}}
\end{center}
\caption{Empirical female ratio ${\P}(\bd)$ 
and multi-task FNN estimated female ratio $\widehat{\P}(\bd)$
under the chosen drop-out rates.}
\label{female ratio}
\end{table}

We use these data sets with drop-outs (incomplete protected
information) to perform two different model fittings.
Firstly, in a more naive approach, we just fit a plain-vanilla FNN \eqref{plain-vanilla FNN},
such that we only use those observations for which the discriminatory information is available
and we drop all insurance policies with incomplete information.
Thus, if e.g.~the drop-out probability is 80\% we only use the remaining 20\% of the data for 
model fitting, for which the discriminatory information $\bD_i$ is available.
Secondly, this naive approach is challenged
by a multi-task FNN \eqref{multi-task learning} fitted with the loss function \eqref{multi-task loss NA}, which accounts for partial availability of discriminatory information, but uses
the entire portfolio for model fitting.

\begin{figure}[htb!] 
\begin{center}
\begin{minipage}{0.6\textwidth}
\begin{center}
\includegraphics[width=\linewidth]{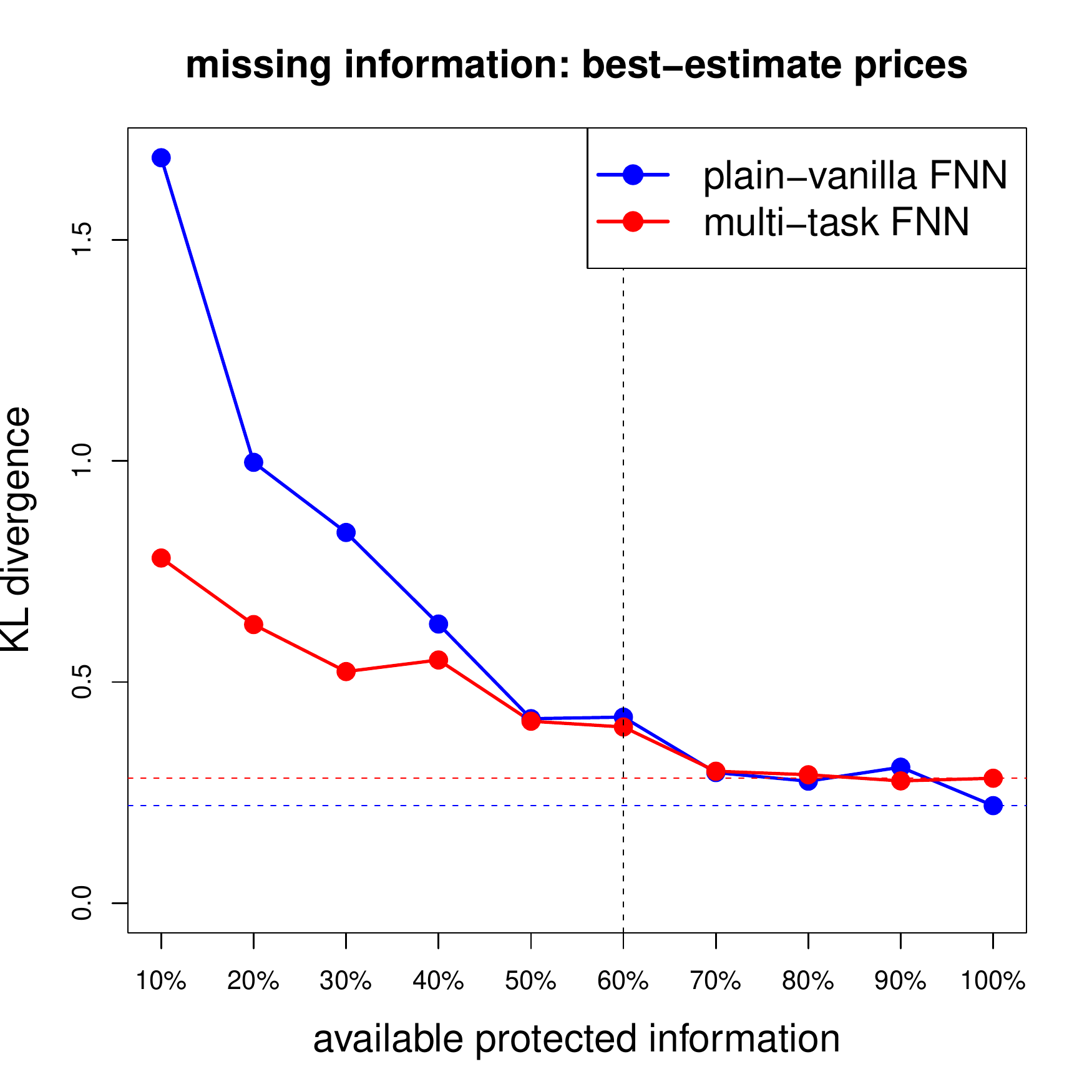}
\end{center}
\end{minipage}
\end{center}
\caption{Comparison of the (naive) plain-vanilla FNN and the multi-task FNN under missing
  discriminatory information: KL divergences from the fitted best-estimate prices $\mu(\bx,\bd)$
  to the true best-estimates $\lambda(\bx,\bd)$; scale on $y$-axis is in $10^{-3}$.}
\label{Figure: MCAR1}
\end{figure}

The results are presented in Figure \ref{Figure: MCAR1}. This figure shows the 
KL divergences from the fitted best-estimate prices $\mu(\bx,\bd)$ to the true best-estimate
price $\lambda(\bx,\bd)$;
the dotted horizontal lines illustrate the results of Table \ref{model accuracy} reflecting the case of full discriminatory information.
We observe that if sufficient discriminatory information $\bD_i$ is available, then we have
a similar performance between the plain-vanilla
FNN and the multi-task FNN. However, below a critical amount of discriminatory information (60\% in our case; vertical dotted line)
we give clear preference to the multi-task FNN because its KL divergence from the true model is clearly smaller,
i.e., we receive a more accurate model from the multi-task FNN.

\begin{figure}[htb!] 
\begin{center}
\begin{minipage}{0.6\textwidth}
\begin{center}
\includegraphics[width=\linewidth]{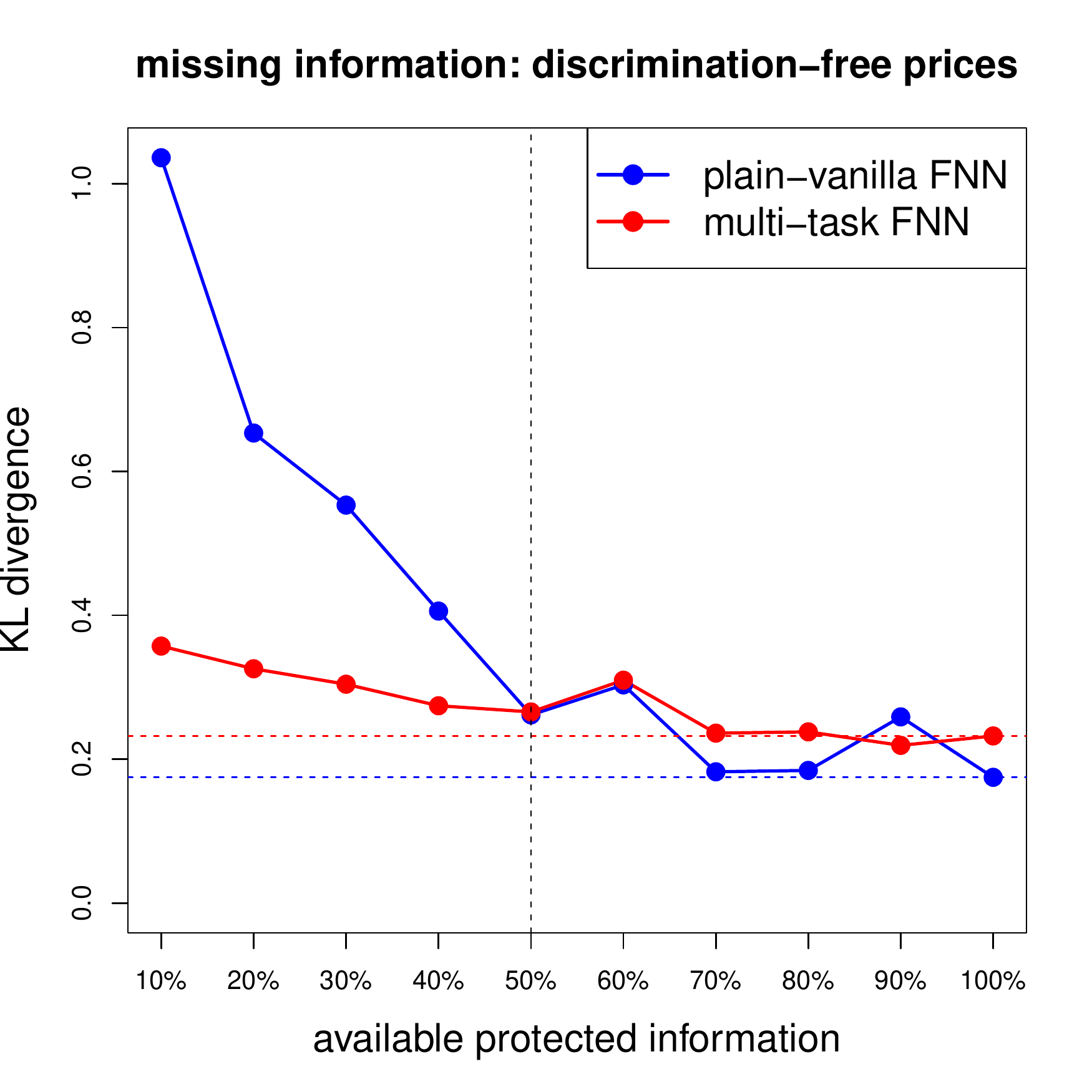}
\end{center}
\end{minipage}
\end{center}
\caption{Comparison of the (naive) plain-vanilla FNN and the multi-task FNN under missing
  discriminatory information: KL divergences from the fitted discrimination-free 
  insurance prices $\mu^*(\bx)$
  to the true discrimination-free insurance prices $\lambda^*(\bx)$; scale on $y$-axis is in $10^{-3}$.}
\label{Figure: MCAR2}
\end{figure}

This preference for the multi-task FNN carries over to the discrimination-free insurance 
prices. In Figure \ref{Figure: MCAR2} we illustrate the KL divergences
from the estimated discrimination-free insurance prices $\mu^*(\bx)$ to the true discrimination-free
insurance price $\lambda^*(\bx)$; the case of full discriminatory information is denoted by the
horizontal dotted lines and corresponds to rows (b2) and (d2) of Table 
\ref{model accuracy 3}. We observe smaller KL divergences of the multi-task FNN
approach if we have discriminatory information on less than 50\% of the policies (vertical
black dotted line). Furthermore, we notice that in the multi-task FNN case, the KL divergence deteriorates only mildly, when the availability of discriminatory information decreases beyond the 50\% point.

The discrimination-free insurance prices of Figure \ref{Figure: MCAR2} have simply used the empirical estimates for $\P(\bd)$  from the
insurance policies where full information is available,  see
row `empirical $\P(\bd)$' of Table \ref{female ratio}. 
Having the fitted multi-task FNN we can also
use the estimated categorical probabilities $p_k(\bX_i)$, $1\le k \le K$, 
from \eqref{multi-task learning} to estimate the
distribution of the protected covariates. Namely, we get an estimate 
\begin{equation}\label{multi-task FNN gender estimation}
\widehat{\P}(\bd_k) = \frac{1}{n} \sum_{i=1}^n p_k(\bX_i) \qquad \text{ for $1\le k \le K$.}
\end{equation}
These estimates (in the binary gender case $K=2$)
are presented on the last row of Table \ref{female ratio}.
We observe that they match the empirical estimates, and only for high drop-out
probabilities we have some differences to the empirical estimates. The
specific choice only has a marginal influence on the prices.

\subsubsection{Not missing completely at random}
In the last example, illustrated by Figures  \ref{Figure: MCAR1} and \ref{Figure: MCAR2}, discriminatory information
$\bD_i$ was removed completely at random, i.e., 
$\bD_i$ was set to {\tt NA} by an i.i.d.~Bernoulli random variable with a fixed 
drop-out rate. 
However, it might be that the missingness of protected information is not completely
independent from the remaining covariates $\bX_i$. Of course, there are
many different ways in which this could happen, and we just provide here one particular
example. We take as a baseline the i.i.d.~Bernoulli case $\bD_i={\tt NA}$ with a drop-out probability of 70\%. We then modify this case by choosing a higher drop-out rate
for the gender information on the policies
${\mathcal M}=\{ X_1 \le 45, X_2 ={\rm smoker}\}$. Besides the base case of
70\%, we choose the drop-out rates of 80\% and 90\% on ${\mathcal M}$. 
Note that ${\mathcal M}$ collects the smokers with ages below 45, 
and by the choice of our population distribution, 79.9\% on this sub-portfolio
are female. 

\begin{table}[htb]
\begin{center}
{\small
\begin{tabular}{|l|c|cc|}
\hline 
 & missing at random & \multicolumn{2}{c|}{not missing at random}\\\hline
 drop-out rate on ${\mathcal M}$ & 70\%  & ~~80\%  & 90\%\\
\hline
overall drop-out rate & 70\% & ~~78\% & 86\% \\  
\hline
empirical ${\P}(\bd)$ & 44.3\%& 42.4\%& 40.3\%\\
multi-task $\widehat{\P}(\bd)$ & 44.3\%& 45.0\%& 44.9\%\\
  \hline

\end{tabular}}
\end{center}
\caption{Not completely missing at random: resulting overall drop-out rates, and
empirical female ratios $\P(\bd)$ and multi-task FNN estimated female ratio $\widehat{\P}(\bd)$.}
\label{NCAR 0}
\end{table}

Table \ref{NCAR 0} shows the resulting overall drop-out rates.
These drop-out rates are no longer missing completely at random, because we have higher
drop-out rates on ${\mathcal M}$. In our case this results in a biased empirical gender estimate.
This can be seen from the row `empirical $\P(\bd)$' which simply calculates the empirical female
ratio on the policies where full information is available. Having the fitted multi-task FNN, we can also estimate the female
ratio using the categorical probability estimates of $p_k(\bX_i)$, see \eqref{multi-task FNN gender estimation}. This provides us with the results on the last row of Table \ref{NCAR 0}.
We observe that these estimates are close to unbiased, the true value being 45\%.
We can use this multi-task FNN to estimate $\widehat{\P}(\bd)$ as pricing measure for the discrimination-free
insurance price in the case of data not missing completely at random.

\begin{table}[htb]
\begin{center}
{\footnotesize
\begin{tabular}{|l||c|cc|}
\hline 
 & \multicolumn{3}{|c|}{KL divergence \eqref{KL divergence} to $\lambda^*(\bx)$}\\
 \hline 
 & missing at random & \multicolumn{2}{c|}{not missing at random}\\
 \multicolumn{1}{|r||}{drop-out rate on ${\mathcal M}$} & 70\%  & 80\%  & 90\%\\
\hline\hline
  (b3) plain-vanilla FNN discrimination-free & 0.5532 &0.8153& 0.9306\\
 (d3)  multi-task FNN discrimination-free  & 0.3042 &0.3698& 0.3750\\
  
\hline

\end{tabular}}
\end{center}
\caption{Model accuracy of the discrimination-free insurance prices
  $\mu^*(\bx)$ relative to the true discrimination-free insurance price $\lambda^*(\bx)$
  where the drop-out probability of the gender information is not missing completely at random;
  the KL divergences are stated in $10^{-3}$; the resulting overall drop-out rates are given in 
  Table \ref{NCAR 0}.}
\label{model accuracy 4}
\end{table}

\begin{figure}[htb!] 
\begin{center}
\begin{minipage}{0.6\textwidth}
\begin{center}
\includegraphics[width=\linewidth]{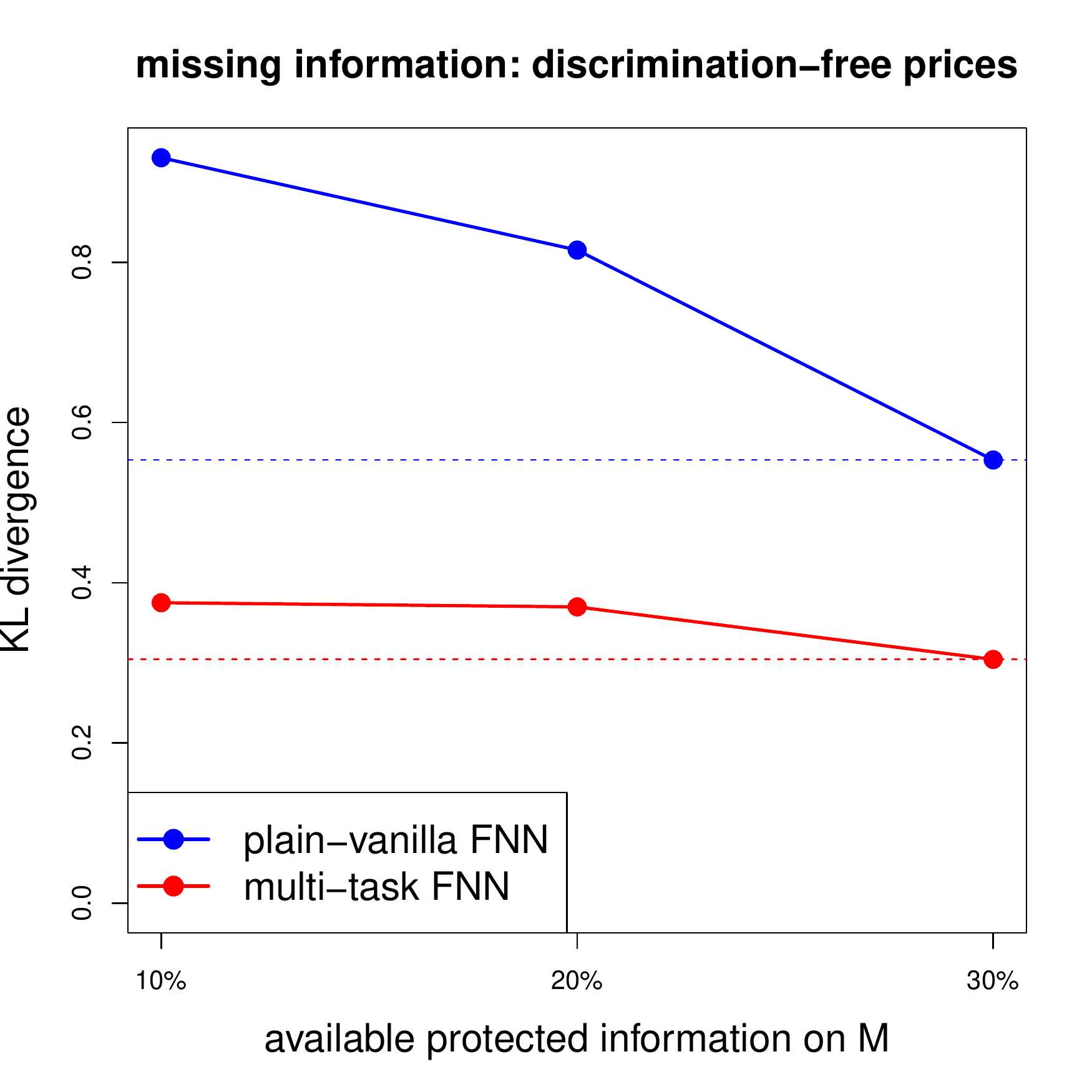}
\end{center}
\end{minipage}
\end{center}
\caption{Comparison of the (naive) plain-vanilla FNN and the multi-task FNN where
the drop-out probability is not missing completely at random: KL divergences from the fitted discrimination-free insurance prices $\mu^*(\bx)$
  to the true discrimination-free insurance prices $\lambda^*(\bx)$; 
  scale on $y$-axis is in $10^{-3}$; the $x$-axis gives the available
  discriminatory information $\bD_i$ on ${\mathcal M}$ and 
  the resulting overall drop-out rates are given in 
  Table \ref{NCAR 0}.}
\label{Figure: NMCAR3}
\end{figure}

Table \ref{model accuracy 4} and Figure \ref{Figure: NMCAR3} present the results.
The base case is the i.i.d.~Bernoulli drop-out case with a drop-out rate of 70\%, taken from Figure \ref{Figure: MCAR2}. This base case
is modified to a higher drop-out rate of 80\% and 90\%, respectively, on sub-portfolio ${\mathcal M}$; see Table \ref{NCAR 0} for the resulting overall drop-out rates.
For the pricing measure $\P^*(\bd)$ we choose the empirical probability $\P(\bd)$
in the plain-vanilla FNN case and the multi-task FNN estimate $\widehat{\P}(\bd)$ in the
multi-task FNN case, see Table \ref{NCAR 0}. At the first sight, this does not seem to be an
entirely fair comparison because the former estimates are biased. However, there is no simple
way in the plain-vanilla FNN case to receive better gender estimates, whereas in the multi-task
FNN case we obtain these better estimates as an integral part of the prediction model.
In this not missing completely at random example we arrive at the the same conclusion, namely, that the multi-task FNN shows superior performance. We remark that even if we would use the same biased gender estimates
also in the multi-task FNN to calculate the discrimination-free insurance prices we would
come to the same conclusion.

\begin{remark}
In Remark \ref{demographic discrimination} we have discussed that  discrimination may also result from the
fact that a certain sub-population is under-represented in the data and, hence,
we may have a poorly fitted model on this part of the covariate space. This form of
discrimination is directly related to incomplete data not missing completely at random, which
is relevant when fitting the best-estimate price $\mu(\bx,\bd)$. The present section
has shown that the multi-task FNN can help to improve model accuracy if under-representation
is caused by missing discriminatory information. However, if the sub-population is under-represented
per se, e.g., there are only few elderly female smokers in the portfolio, then this multi-task FNN
cannot resolve the fundamental class imbalance problem.
\end{remark}


\section{Concluding remarks}
\label{Conclusions}
Addressing the problem of indirect or proxy discrimination involves an apparent paradox: in order to compensate for the potentially discriminatory effect of implicitly inferring policyholders' protected characteristics, information on these very characteristics must be available for regression modeling. Resolving this tension poses clear legal, regulatory and technical challenges. Here, focusing on the latter, we provided a multi-task neural network learning framework, which can generate insurance prices that are free from indirect discrimination. We demonstrated that this multi-task architecture is competitive to conventional approaches when full information is available, while clearly outperforming them in the case of partial information. 

Nonetheless, there is an aspect of the technical challenge that we have not yet fully addressed. Practically, we still need discriminatory information $\bD$ for a part of the 
portfolio in order to fit our model. Hence, a scheme needs to be in place that allows insurers to access such protected information for a subset of policies.
Such a scheme may be constructed commercially, e.g., by offering special
discounts to customers who are willing to disclose information on protected characteristics. Besides addressing privacy concerns, a difficulty
with such an approach is to ensure (or mitigate) the potential selection bias that such a commercial
promotion will generally have. While our case study illustrated good performance of our model when data are not missing at random, further work on this topic is required.

A related issue is that, to go from best-estimate prices to discrimination-free 
insurance prices, we need to choose the pricing measure $\P^*(\bd)$. A natural candidate is to use the empirical version of $\P(\bd)$, but since this choice will be based on a subset of the portfolio, the question again arises as to whether this subset is representative of the entire portfolio. 
In the multi-task FNN we receive an estimate for $\P(\bd)$ as
an integral part of the prediction model by the averaging in \eqref{multi-task FNN gender estimation} of
the estimated categorical probabilities $p_k(\bX_i)$.
Alternatively, techniques from survey sampling could be used in order to obtain an estimate of $\P(\bd)$, using so-called indirect questioning. These techniques were constructed in order to obtain unbiased estimates of population proportions of a single sensitive dichotomous characteristic, such as drug use and sexual preference, based on open answer questionnaires, see the seminal paper by Warner \cite{warner1965randomized}.
For more general categorical sensitive characteristics, alternative techniques can be used;
see Lager{\aa}s--Lindholm \cite{lageraas2020ask} and the survey of Chaudhuri--Christofides \cite{chaudhuri2013indirect}. Regardless of the specific technique employed, 
by obtaining a suitable total population estimate for $\P(\bd)$ it is possible to assess whether 
the sub-portfolio has been sampled with data missing completely at random or not.


\end{document}